\documentclass[lettersize,journal]{IEEEtran}
\usepackage{amsmath,amsfonts}
\usepackage{algorithmic}
\usepackage{algorithm}
\usepackage{array}
\usepackage{textcomp}
\usepackage{stfloats}
\usepackage{url}
\usepackage{verbatim}
\usepackage{graphicx}
\usepackage{cite}
\usepackage{tikz}
\usepackage[breaklinks,colorlinks,linkcolor=blue,citecolor=blue,urlcolor=blue]{hyperref}
\usepackage{simpleicons}
\usepackage{pifont}
\usepackage{orcidlink}
\usepackage{bm}
\usepackage{listings}
\usepackage{xcolor}
\usepackage{tcolorbox}
\usepackage{makecell}
\usepackage{pifont}

\usepackage{booktabs}
\usepackage{amssymb}
\usepackage{xcolor}
\usepackage{multirow}

\usepackage{color, colortbl}
\definecolor{Gray}{gray}{0.9}

\newcolumntype{S}{>{\footnotesize}c}

\newcommand{\DIORRRSVGStatistics}{
    \begin{table}[]
        \centering
        \renewcommand\arraystretch{1.1}
        \caption{Comparison of existing remote sensing visual grounding datasets and our proposed \textbf{DIOR-R-RSVG}.}
        \resizebox{\linewidth}{!}
        {
            \begin{tabular}{ccccccc}
              \toprule
              \textbf{Dataset} & \textbf{Year} & \textbf{Train} & \textbf{Val} & \textbf{Test} & \textbf{Image} & \textbf{Size} \\
              \midrule
              DIOR-RSVG~\cite{diorrsvg} & 2023 & 26,991 & 3,829 & 7,500 & 17,402 &(800,800) \\
              VRSBench~\cite{vrsbench}  & 2024 & 36,313 & - & 16,159 & 29,614 &(512,512) \\
              AVVG~\cite{refgeo}        & 2025 & 26,482 & - & 6,139 & 624 &(4000,2250) \\
              \rowcolor{gray!10} DIOR-R-RSVG               & 2026 & 26,991 & 3,829 & 7,500 & 17,402 &(800,800) \\
              \bottomrule
            \end{tabular}
        }
        \label{Tab:DIORRRSVGStatistics}
    \end{table}
}

\newcommand{\OtwoVGTransResult}{
    \begin{table*}[]
        \centering
        \renewcommand\arraystretch{1.1}
        \caption{Results of \textbf{O$^2$-VG-Trans} on the DIOR-R-RSVG, VRSBench, and AVVG datasets. The \textbf{best} results are highlighted in \textbf{bold}.}
        \resizebox{\linewidth}{!}
        {
            \begin{tabular}{lcccccccccc}
              \toprule
              \textbf{Method} & \makecell{\textbf{Image} \\ \textbf{Backbone}} & \makecell{\textbf{Text} \\ \textbf{Backbone}} & \textbf{Params} 
    & \textbf{Pr@0.5} & \textbf{Pr@0.6} & \textbf{Pr@0.7} & \textbf{Pr@0.8} & \textbf{Pr@0.9} & \textbf{meanIoU} & \textbf{cumIoU} \\
              \midrule
              \multicolumn{11}{c}{\textit{VRSBench}} \\
              Rotated GiT-B~\cite{git} & SAM ViT & - & 159M & 47.23 & 35.86 & 22.90 & 10.56 & 2.20 & 41.90 & 51.02 \\
              Rotated GDINO~\cite{dino} & ResNet50 & BERT & 172M & 58.38 & 49.91 & 33.02 & 15.46 & 3.02 & 45.73 & 54.28 \\
              Rotated EGDINO~\cite{effgrounddino} & ResNet50 & BERT & 169M & 61.57 & 52.26& 36.81 & 17.74 & 3.68 & 48.34 & 56.97 \\
              \rowcolor{gray!10} \textbf{O$^2$-VG-Trans (ours)} & ResNet50 & BERT & 172M & \textbf{67.71} & \textbf{60.13} & \textbf{46.53} & \textbf{27.11} & \textbf{7.63} & \textbf{55.01} & \textbf{61.14} \\
              \midrule
              \multicolumn{11}{c}{\textit{AVVG}} \\
              Rotated GiT-B~\cite{git} & SAM ViT & - & 159M & 2.37 & 1.23 & 0.30 & 0.00 & 0.00 & 1.41 & 2.05 \\
              Rotated GDINO~\cite{dino} & ResNet50 & BERT & 172M & 12.64 & 10.73 & 8.82 & 2.91 & 0.46 & 9.52 & 11.74 \\
              Rotated EGDINO~\cite{effgrounddino} & ResNet50 & BERT & 169M & 14.42 & 12.26 & 10.35 & 4.54 & 0.53 & 10.38 & 12.21 \\
              \rowcolor{gray!10} \textbf{O$^2$-VG-Trans (ours)} & ResNet50 & BERT & 172M & \textbf{18.00} & \textbf{17.04} & \textbf{14.37} & \textbf{7.83} & \textbf{0.98} & \textbf{14.59} & \textbf{16.64} \\
              \midrule
              \multicolumn{11}{c}{\textit{DIOR-R-RSVG}} \\
              Rotated GiT-B~\cite{git} & SAM ViT & - & 159M & 43.52 & 35.93 & 27.04 & 15.59 & 3.49 & 38.64 & 55.41 \\
              Rotated GDINO~\cite{dino} & ResNet50 & BERT & 172M & 56.84 & 48.16 & 38.28 & 25.63 & 8.72 & 47.91 & 58.42 \\
              Rotated EGDINO~\cite{effgrounddino} & ResNet50 & BERT & 169M & 60.37 & 52.84 & 43.16 & 29.87 & 11.94 & 50.87 & 61.73 \\
              \rowcolor{gray!10} \textbf{O$^2$-VG-Trans (ours)} & ResNet50 & BERT & 172M & \textbf{67.23} & \textbf{62.00} & \textbf{54.12} & \textbf{39.47} & \textbf{17.01} & \textbf{56.73} & \textbf{67.33} \\
              \bottomrule
            \end{tabular}
        }
        \label{Tab:OtwoVGTransResult}
    \end{table*}
}

\newcommand{\OtwoVGUniRecall}{
    \begin{table*}[]
        \centering
        \renewcommand\arraystretch{1.1}
        \caption{Results of \textbf{O$^2$-VG-Uni} on the DIOR-R-RSVG, VRSBench, and AVVG datasets for universal oriented proposals.}
        \resizebox{\linewidth}{!}
        {
            \begin{tabular}{l|lcc|ccc|ccc|ccc}
              \toprule
                     &   \textbf{Image}  &        &     & \multicolumn{3}{c|}{\textbf{VRSBench}}  & \multicolumn{3}{c|}{\textbf{AVVG}}  & \multicolumn{3}{c}{\textbf{DIOR-R-RSVG}} \\
              \textbf{Method} & \textbf{Backbone} & \textbf{Params} & \textbf{FPS} & \textbf{AR}$_{50}$     & \textbf{AR}$_{75}$  & \textbf{AR}$_{50:95}$  & \textbf{AR}$_{50}$    & \textbf{AR}$_{75}$ & \textbf{AR}$_{50:95}$   & \textbf{AR}$_{50}$    & \textbf{AR}$_{75}$  & \textbf{AR}$_{50:95}$    \\
              \midrule
              \textit{100 proposals} &      & & & & & &       & & &       & \\
              Rotated RPN~\cite{faster_rcnn} &   R18 & 14M & 23 & 83.14 & 40.78 & 43.81 & 54.98 & 14.86 & 23.56 &  85.08  & 50.36 & 48.90 \\
              Rotated RPN~\cite{faster_rcnn} &  R50 & 27M & 22 &  85.79 & 49.03 & 48.81 & 60.35 & 19.02 &27.00 &  \textbf{88.21} & 54.65 & 52.18 \\
              \rowcolor{gray!10} \textbf{O$^2$-VG-Uni (ours)} & R50vd & 44M/512 & \textbf{83} & \textbf{86.79} & \textbf{58.48} & \textbf{55.48} & \textbf{97.28} & \textbf{86.61} & \textbf{73.61} & 86.23 & \textbf{64.41} & \textbf{59.76} \\
              \midrule
              300 proposals &      & & & & & &  & & &       &   \\
              Rotated RPN~\cite{faster_rcnn} & R18 & 14M & 23 & 87.77 & 45.36 & 47.50 & 80.65   & 23.98 & 36.29 & 89.80 & 54.69 & 52.38 \\
              Rotated RPN~\cite{faster_rcnn} & R50 & 27M & 22 & \textbf{89.78} & 53.72 & 52.07 &   86.47    & 30.01 &40.67 & \textbf{91.72} & 58.31 & 54.98 \\  
              \rowcolor{gray!10} \textbf{O$^2$-VG-Uni (ours)} & R50vd & 44M/512 & \textbf{83} & 87.89 & \textbf{59.60} & \textbf{56.41} & \textbf{98.35} & \textbf{88.52} & \textbf{75.36} & 88.53 & \textbf{65.20} & \textbf{60.78} \\
              \bottomrule
            \end{tabular}
        }
        \label{Tab:OtwoVGUniRecall}
    \end{table*}
}

\newcommand{\OtwoVGUniRetrieval}{
    \begin{table}[]
        \centering
        \renewcommand\arraystretch{1.1}
        \caption{Results of \textbf{O$^2$-VG-Uni}on the DIOR-R-RSVG, VRSBench, and AVVG datasets for the object retrieval task.}
        \resizebox{\linewidth}{!}
        {
            \begin{tabular}{lcc|ccc}
              \toprule
              \textbf{Method} & \textbf{Dataset} & \textbf{thre.} & \textbf{P}     & \textbf{R}  & \textbf{F1}    \\
              \midrule
              Clip~\cite{clip}       & VRSBench  & 0.3   & 36.77 & 5.41  &7.32   \\
              RemoteClip~\cite{remoteclip} & VRSBench & 0.3 & \textbf{76.72} &  30.34 &  37.69     \\
              \rowcolor{gray!10} \textbf{O2-VG-Uni (ours)}  & VRSBench & 0.3 & 70.31  &\textbf{53.11} & \textbf{56.86}     \\
                            \midrule
              Clip~\cite{clip}       & AVVG  & 0.3 & 0.00  &  0.00 & 0.00  \\
              RemoteClip~\cite{remoteclip} & AVVG & 0.3  & 0.00  &  0.00 & 0.00  \\
              \rowcolor{gray!10} \textbf{O2-VG-Uni (ours)} & AVVG & 0.3 & \textbf{17.38}  &  \textbf{2.45} & \textbf{3.65}    \\
              \midrule
              Clip~\cite{clip}       & DIOR-R-RSVG  & 0.3  & 26.62  & 6.06 & 8.11   \\
              RemoteClip~\cite{remoteclip} & DIOR-R-RSVG & 0.3  & \textbf{60.62} &  17.65 &  23.33    \\
              \rowcolor{gray!10} \textbf{O2-VG-Uni (ours)} & DIOR-R-RSVG & 0.3 & 59.62  &  \textbf{55.26} & \textbf{49.25}     \\
              \bottomrule
            \end{tabular}
        }
        \label{Tab:OtwoVGUniRetrieval}
    \end{table}
}

\newcommand{\OtwoVGVLMResult}{
    \begin{table*}[]
        \centering
        \renewcommand\arraystretch{1.1}
        \caption{Comparison of the results of the \textbf{O$^2$-VG-VLM} and other methods on the DIOR-R-RSVG, VRSBench, and AVVG datasets.}
        \resizebox{\linewidth}{!}
        {
            \begin{tabular}{lccccccccccc}
              \toprule
              \textbf{Method} & \makecell{\textbf{Vision} \\ \textbf{Encoder}} & \textbf{Decoder} & \textbf{\#P} & \textbf{Pr@0.5} & \textbf{Pr@0.6} & \textbf{Pr@0.7} & \textbf{Pr@0.8} & \textbf{Pr@0.9} & \textbf{meanIoU} & \textbf{cumIoU} & \textbf{BPS} \\
              \midrule
              \multicolumn{11}{c}{\textit{VRSBench}} \\
              GeoChat w/o ft~\cite{geochat} & Clip ViT & Vicuna1.5 & 7B & 13.71 & 7.23 & 2.72 & 0.74 & 0.04 & 23.14 & 27.79 & 1\\
              GeoChat~\cite{geochat} & Clip ViT & Vicuna1.5 & 7B & 25.84 & 16.32 & 8.47 & 2.91 & 0.46 & 32.18 & 37.65   & 1 \\
              GeoGround~\cite{geoground} & Clip ViT & Vicuna1.5 & 7B & 35.76 & 25.48 & 15.23 & 6.87 & 1.12 & 39.74 & 45.83  & 1\\
              \rowcolor{gray!10} \textbf{O$^2$-VG-VLM (ours)} & Moon ViT & Qwen2.5 & 3B & \textbf{69.67} & \textbf{61.69} & \textbf{50.11} & \textbf{33.38} & \textbf{13.04} & \textbf{59.84} & \textbf{59.36} & \textbf{12}\\
              \midrule
              \multicolumn{11}{c}{\textit{AVVG}} \\
              GeoChat w/o ft~\cite{geochat} & Clip ViT & Vicuna1.5 & 7B & 0.02 & 0.00 & 0.00 & 0.00 & 0.00 & 1.06 & 1.87  & 1\\
              GeoChat~\cite{geochat} & Clip ViT & Vicuna1.5 & 7B & 12.48 & 9.76 & 7.84 & 2.52 & 0.42 & 9.31 & 10.86  & 1 \\
              GeoGround~\cite{geoground} & Clip ViT & Vicuna1.5 & 7B & 13.73 & 11.92 & 8.96 & 3.73 & 0.50 & 10.08 & 11.94  & 1\\
              \rowcolor{gray!10} \textbf{O$^2$-VG-VLM (ours)} & Moon ViT & Qwen2.5 & 3B & \textbf{31.29} & \textbf{29.24} & \textbf{25.47} & \textbf{17.81} & \textbf{4.00} & \textbf{26.44} & \textbf{27.52} & \textbf{12} \\
              \midrule
              \multicolumn{11}{c}{\textit{DIOR-R-RSVG}} \\
              GeoChat w/o ft~\cite{geochat} & Clip ViT & Vicuna1.5 & 7B & 25.76 & 15.89 & 6.57 & 1.53 & 0.13 & 28.74 & 43.52  & 1\\
              GeoChat~\cite{geochat} & Clip ViT & Vicuna1.5 & 7B & 50.47 & 42.31 & 29.82 & 17.48 & 5.91 & 44.63 & 56.24  & 1\\
              GeoGround~\cite{geoground} & Clip ViT & Vicuna1.5 & 7B & 58.28 & 50.71 & 40.16 & 26.84 & 9.73  & 48.21 & 59.37 & 1 \\
              \rowcolor{gray!10} \textbf{O$^2$-VG-VLM (ours)} & Moon ViT & Qwen2.5 & 3B & \textbf{78.35} & \textbf{73.49} & \textbf{65.47} & \textbf{50.74} & \textbf{24.48} & \textbf{67.85} & \textbf{74.66}  & \textbf{12}\\
              \bottomrule
            \end{tabular}
        }
        \label{Tab:OtwoVGVLMResult}
    \end{table*}
}

\newcommand{\OtwoVGTransEachModule}{
    \begin{table}[!t]
        \centering
        \caption{Ablation study of different components in \textbf{O$^2$-VG-Trans} on the DIOR-R-RSVG dataset.}
        \resizebox{\linewidth}{!}{
        \begin{tabular}{l|cccc}
            \toprule
            \textbf{Method}  & \textbf{Pr@0.5} & \textbf{Pr@0.6} & \textbf{Pr@0.7} & \textbf{meanIoU} \\
            \midrule
            \rowcolor{gray!10} O$^2$-VG-Trans (Full) & 67.23 & 62.00 & 54.12 & 56.73 \\
            \midrule
            \ding{172} w/o encoder fusion       & 62.91 & 56.03 & 48.18 & 51.87 \\
            \ding{173} static query selection   & 64.31 & 58.94 & 50.86 & 53.67 \\
            \ding{174} w/o text cross-attention & 63.27 & 57.82 & 49.36 & 52.61 \\
            \ding{175} word-level text prompt   & 64.02 & 58.61 & 50.47 & 53.42 \\
            \ding{176} axis-aligned points      & 63.84 & 58.37 & 50.11 & 53.18 \\
              \bottomrule
        \end{tabular}
        }
      \label{Tab:OtwoVGTransEachModule}
      \end{table}
}

\newcommand{\OtwoVGUniEachModule}{
    \begin{table}[!t]
        \centering
        \caption{Ablation study of different components in \textbf{O$^2$-VG-Uni} on the DIOR-R-RSVG dataset.}
        \begin{tabular}{l|ccc}
            \toprule
            \textbf{Method}  & \textbf{Param} & \textbf{AR$_{50}$} & \textbf{F1} \\
            \midrule
            \rowcolor{gray!10} O$^2$-VG-Uni (Full)                     & 44M/512 & 88.53  & 49.25  \\
            \midrule
            \ding{172} w/o stage1, train all params in stage2  & 44M/44M & 89.84 & 0.00 \\
            \ding{173} with stage1, train all params in stage2 & 44M/44M & 94.62 & 0.00 \\
            \ding{174} Clip ViT                                & 44M/512 & 83.71 & 42.56\\
            \ding{175} static query selection                  & 44M/512 & 86.27 & 47.18 \\
            \ding{176} axis-aligned points                     & 44M/512 & 85.94 & 45.63  \\
              \bottomrule
        \end{tabular}
      \label{Tab:OtwoVGUniEachModule}
      \end{table}
}

\newcommand{\OtwoVGVLMEachModule}{
    \begin{table}[!t]
        \centering
        \caption{Ablation study of different components in \textbf{O$^2$-VG-VLM} on the DIOR-R-RSVG dataset.}
        \resizebox{\linewidth}{!}{
        \begin{tabular}{l|cccc}
            \toprule
            \textbf{Method}  & \textbf{Pr@0.5} & \textbf{Pr@0.6} & \textbf{Pr@0.7} & \textbf{meanIoU} \\
            \midrule
            \rowcolor{gray!10} O$^2$-VG-VLM (Full)      & 78.35 & 73.49 & 65.47 & 67.85  \\
            \midrule
            \ding{172} tokenize box as text     & 39.84 & 32.17 & 13.36 & 32.72  \\
            \ding{173} four vertex              & 45.83 & 37.26 & 29.14 & 40.62  \\
            \ding{174} w/o universal proposals  & 73.24 & 65.37 & 53.59 & 61.90  \\
            \ding{175} only NTP + slow mode     & 76.62 & 71.71 & 63.82 & 66.02  \\
            \ding{176} only MTP + fast mode     & 74.21 & 68.54 & 60.28 & 62.91  \\
            \bottomrule
        \end{tabular}
        }
      \label{Tab:OtwoVGVLMEachModule}
      \end{table}
}

\newcommand{\OtwoVLMProposal}{
    \begin{table}[]
        \centering
        \renewcommand\arraystretch{1.1}
        \setlength{\tabcolsep}{1.1mm}
        \caption{Effect of universal oriented proposals on \textbf{O$^2$-VG-VLM} across the DIOR-R-RSVG, VRSBench, and AVVG datasets.}
        \resizebox{\linewidth}{!}
        {
            \begin{tabular}{c ccccccc}
                \toprule
                \makecell[c]{\textbf{Universal}\\\textbf{Proposals}}
                    & \textbf{Pr@0.5} & \textbf{Pr@0.6} & \textbf{Pr@0.7} & \textbf{Pr@0.8} & \textbf{Pr@0.9} & \textbf{meanIoU} & \textbf{cumIoU} \\
                \midrule
                \multicolumn{8}{c}{\textit{VRSBench}} \\
                \ding{55}   & 68.38 & 57.64 & 41.57 & 21.51 & 4.77  & 56.82 & 56.75 \\
                \ding{51}   & 69.67 & 61.69 & 50.11 & 33.38 & 13.04 & 59.84 & 59.36 \\
                \midrule
                \multicolumn{8}{c}{\textit{AVVG}} \\
                \ding{55}   & 31.88 & 29.04 & 24.09 & 14.01 & 1.87 & 25.83 & 26.76 \\
                \ding{51}   & 31.29 & 29.24 & 25.47 & 17.81 & 4.00 & 26.44 & 27.52 \\
                \midrule
                \multicolumn{8}{c}{\textit{DIOR-R-RSVG}} \\
                \ding{55}   & 73.24 & 65.37 & 53.59 & 35.82 & 11.42 & 61.90 & 69.33 \\
                \ding{51}   & 78.35 & 73.49 & 65.47 & 50.74 & 24.48 & 67.85 & 74.66 \\
                \bottomrule
            \end{tabular}
        }
        \label{Tab:OtwoVLMProposal}
    \end{table}
}

\newcommand{\OtwoVLMMode}{
    \begin{table}[!t]
        \centering
        \renewcommand\arraystretch{1.1}
        \setlength{\tabcolsep}{1.1mm}
        \caption{Comparison of the slow, fast, and hybrid \textbf{inference modes of O$^2$-VG-VLM} on the DIOR-R-RSVG, VRSBench, and AVVG datasets.}
        \resizebox{\linewidth}{!}
        {
            \begin{tabular}{cccccccc}
              \toprule
              \textbf{Mode} & \textbf{Pr@0.5} & \textbf{Pr@0.6} & \textbf{Pr@0.7} & \textbf{Pr@0.8} & \textbf{Pr@0.9} & \textbf{meanIoU} & \textbf{BPS} \\
              \midrule
              \multicolumn{8}{c}{\textit{VRSBench}} \\
              \textbf{slow}   & 69.91 & 62.26 & 50.45 & 33.60 & 13.32 & 60.09 & 2 \\
              \textbf{fast}   & 69.55 & 61.56 & 49.99 & 33.24 & 12.99 & 59.77 & 13 \\
              \textbf{hybrid} & 69.67 & 61.69 & 50.11 & 33.38 & 13.04 & 59.84 & 12 \\
              \midrule
              \multicolumn{8}{c}{\textit{AVVG}} \\
              \textbf{slow}   & 34.07 & 32.49 & 29.04 & 21.70 & 5.81 & 28.51 & 2 \\
              \textbf{fast}   & 29.64 & 27.49 & 23.69 & 16.25 & 3.48 & 25.04 & 13 \\
              \textbf{hybrid} & 31.29 & 29.24 & 25.47 & 17.81 & 4.00 & 26.44 & 12 \\
              \midrule
              \multicolumn{8}{c}{\textit{DIOR-R-RSVG}} \\
              \textbf{slow} & 79.28 & 74.77 & 66.61 & 51.69 & 25.14 & 68.64 & 2 \\
              \textbf{fast} & 77.95 & 73.06 & 65.16 & 50.55 & 24.44 & 67.57 & 13 \\
              \textbf{hybrid} & 78.35 & 73.49 & 65.47 & 50.74 & 24.48 & 67.85 & 12\\
              \bottomrule
            \end{tabular}
        }
        \label{Tab:OtwoVLMMode}
    \end{table}
}

\usepackage{graphicx}

\newcommand{\teaser}{
    \begin{figure}[!t]
      \centering
      \includegraphics[width=\linewidth]{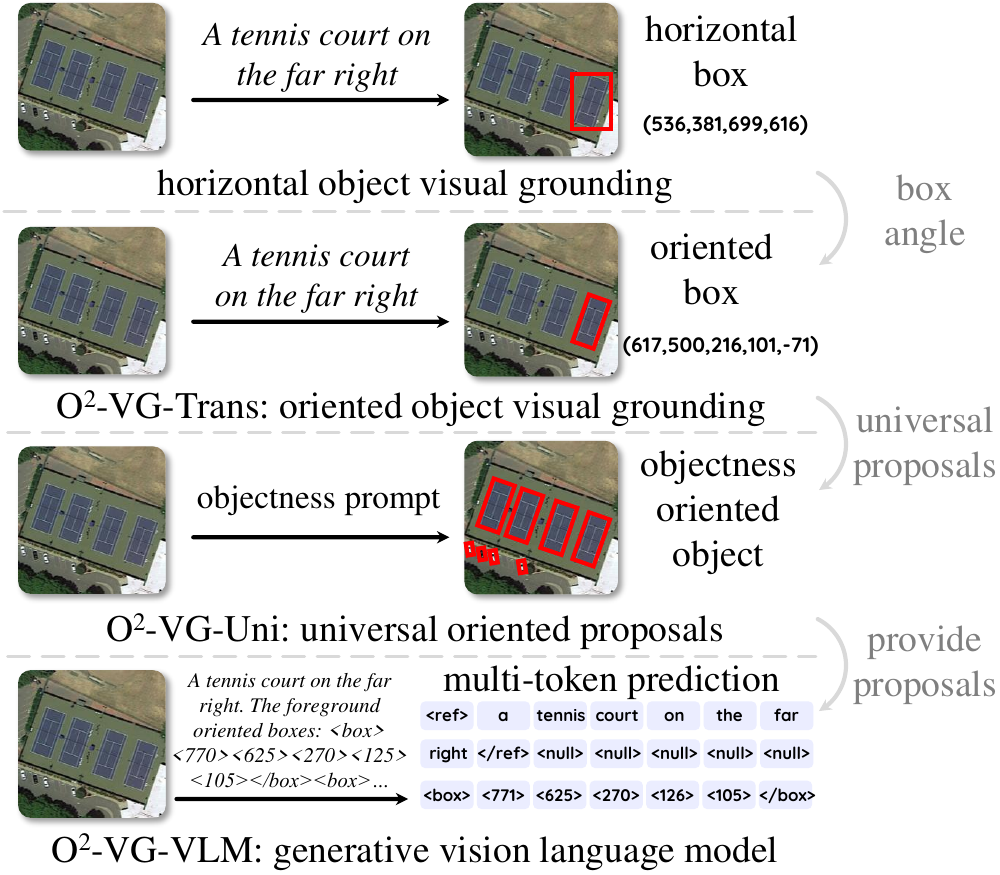}
      \caption{Illustration of oriented object visual grounding in remote sensing images. O$^2$-VG provides a unified family of complementary solutions for oriented object visual grounding.}
      \label{fig:teaser}
    \end{figure}
}

\newcommand{\overview}{
    \begin{figure*}[!t]
    \centering
    \includegraphics[width=0.96\linewidth]{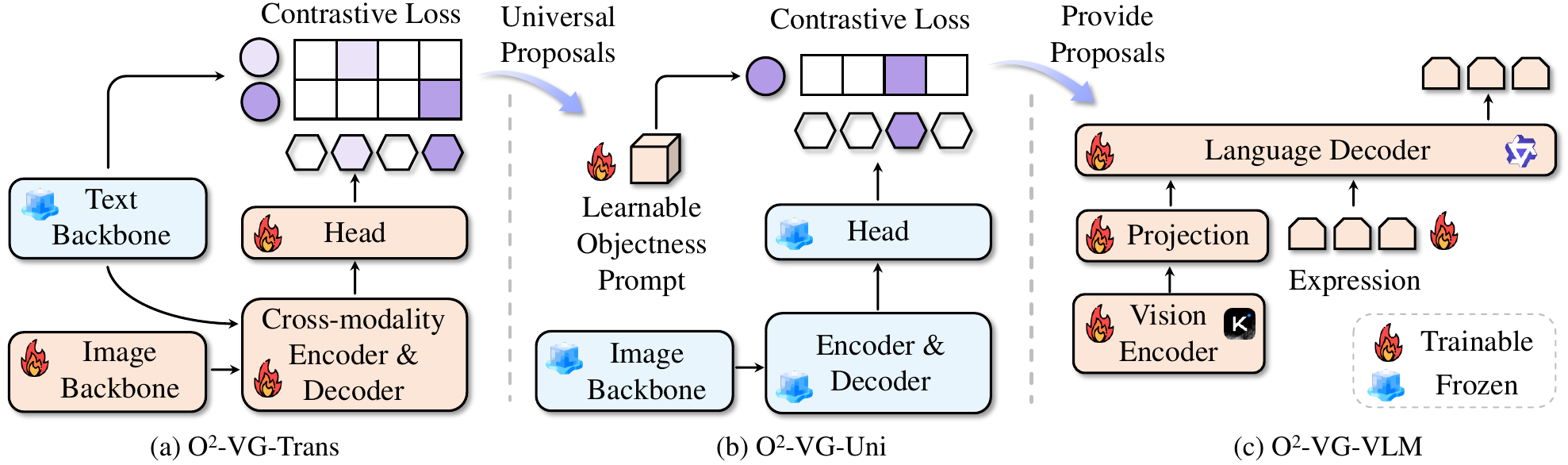}
    \caption{Overview of the O$^2$-VG model family. O$^2$-VG-Trans is an oriented object visual grounding Transformer. O$^2$-VG-Uni provides universal oriented proposals with objectness embeddings and supports object retrieval. O$^2$-VG-VLM is an autoregressive vision-language grounding model that generates oriented box tokens through multi-token prediction within a unified generation framework.}
    \label{fig:overview}
    \end{figure*}

}

\newcommand{\otwovgtrans}{
    \begin{figure*}[!t]
      \centering
      \includegraphics[width=\linewidth]{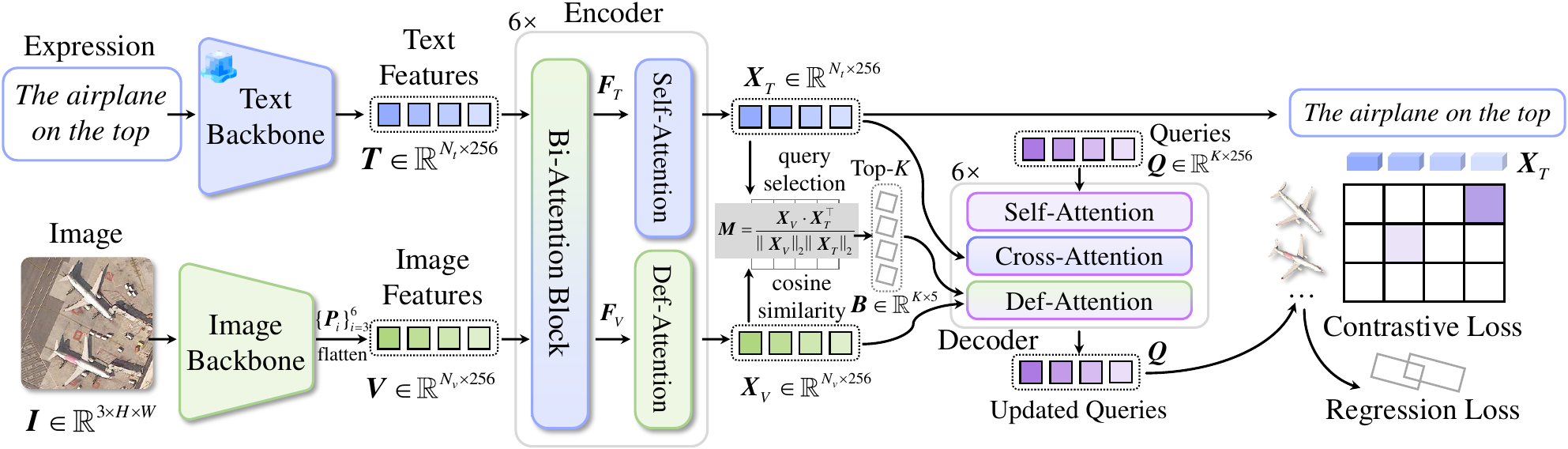}
      \caption{Overview of O$^2$-VG-Trans. O$^2$-VG-Trans is a discriminative cross-modality Transformer for oriented object visual grounding in remote sensing images.}
      \label{fig:o2vgtrans}
    \end{figure*}
}

\newcommand{\otwovguni}{
    \begin{figure*}[!t]
      \centering
      \includegraphics[width=\linewidth]{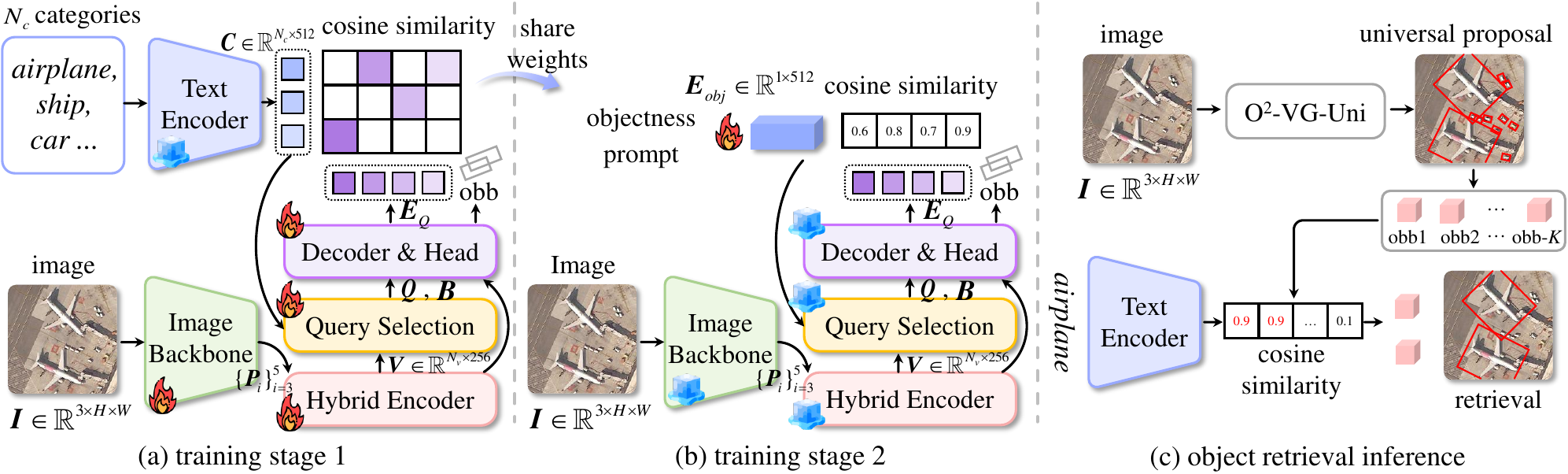}
      \caption{Overview of O$^2$-VG-Uni. (a) Training stage 1 learns text-visual alignment through contrastive learning. (b) Training stage 2 learns universal oriented proposals with objectness prompts. (c) Object retrieval inference with cached proposal embeddings and text queries.}
      \label{fig:o2vguni}
    \end{figure*}
}

\newcommand{\otwovgvlm}{
    \begin{figure*}[!t]
      \centering
      \includegraphics[width=\linewidth]{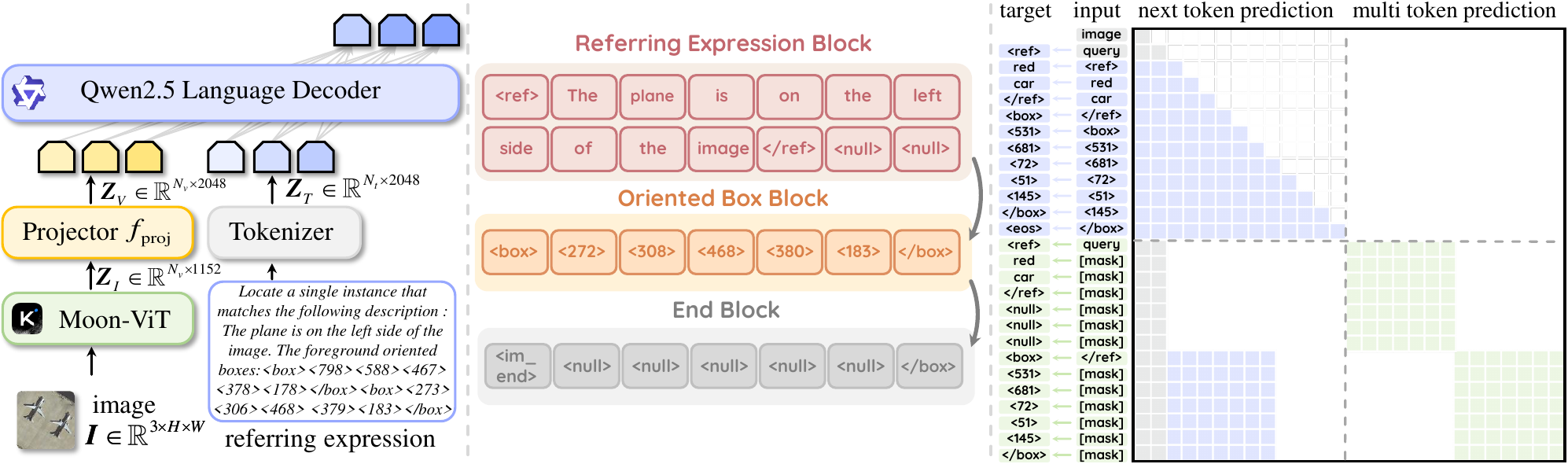}
      \caption{Overview of O$^2$-VG-VLM. Left: the autoregressive vision-language architecture. Middle: the multi-token prediction block. Right: the attention mask for joint NTP-MTP training.}
      \label{fig:o2vgvlm}
    \end{figure*}
}

\newcommand{\diorrrsvgobb}{
    \begin{figure}[!t]
      \centering
      \includegraphics[width=\linewidth]{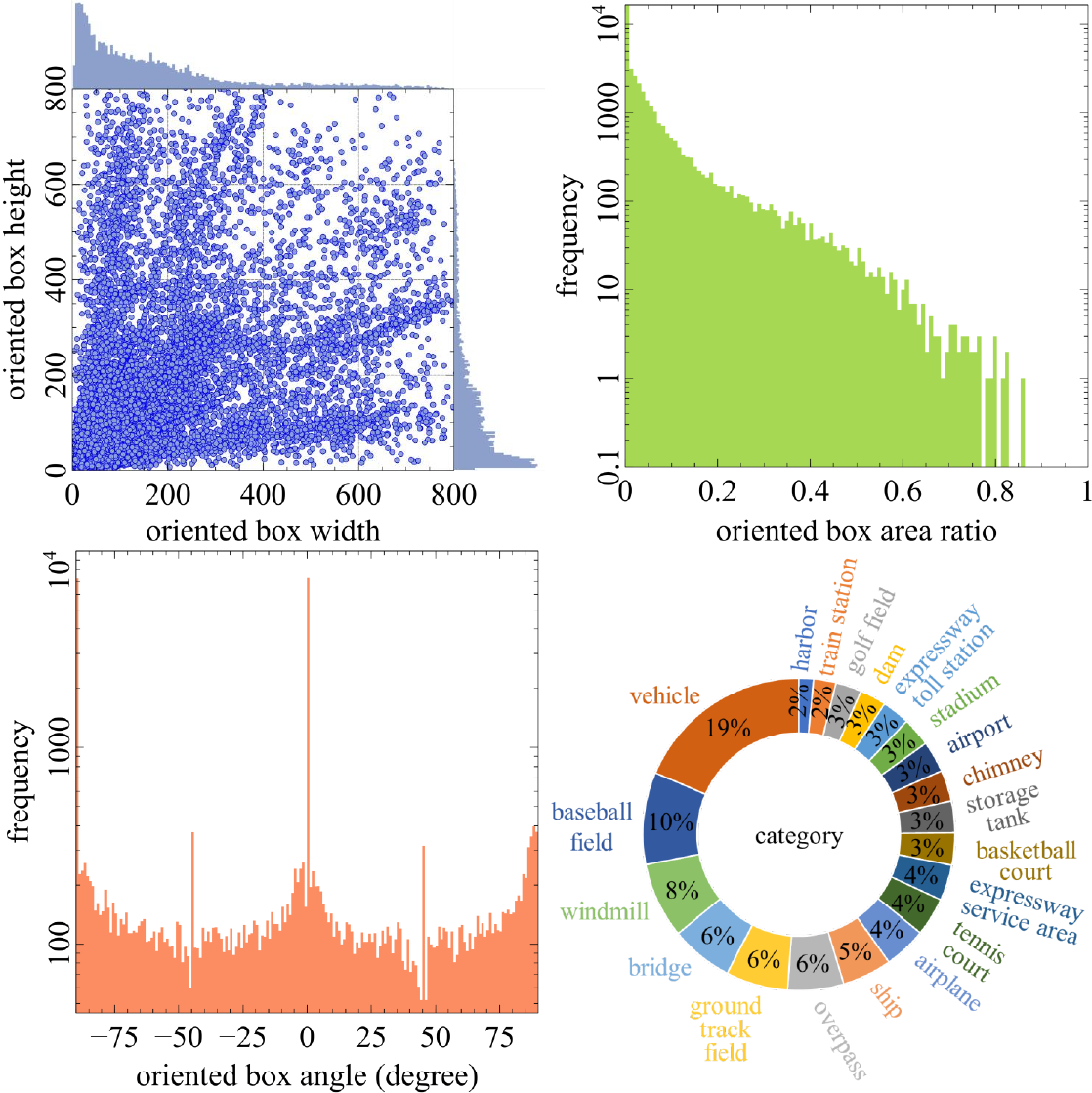}
      \caption{Statistics of oriented object size, area, angle, and categories in the DIOR-R-RSVG dataset.}
      \label{fig:diorrrsvgobb}
    \end{figure}
}

\newcommand{\diorrrsvgexpression}{
    \begin{figure}[!t]
      \centering
      \includegraphics[width=\linewidth]{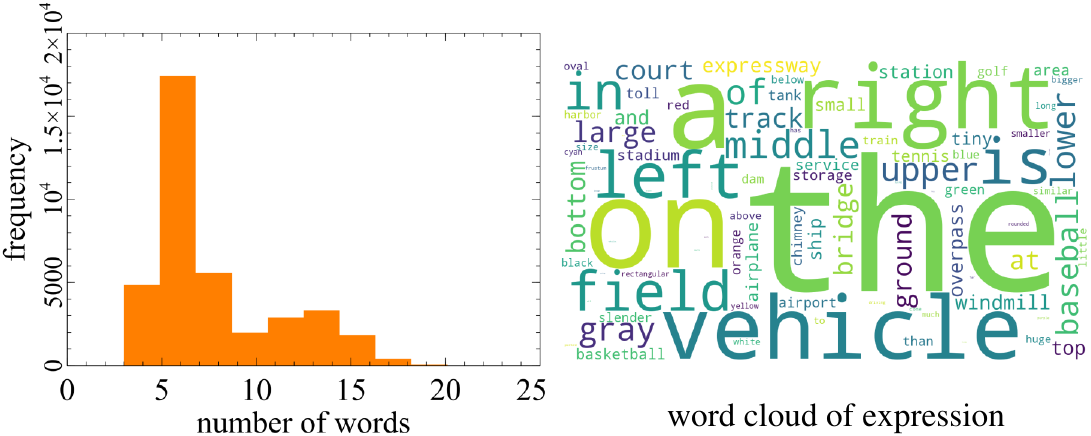}
      \caption{Statistics of expression length and word cloud visualization for the DIOR-R-RSVG dataset.}
      \label{fig:diorrrsvgexpression}
    \end{figure}
}

\newcommand{\diorrrsvghistory}{
    \begin{figure}[!t]
      \centering
      \includegraphics[width=\linewidth]{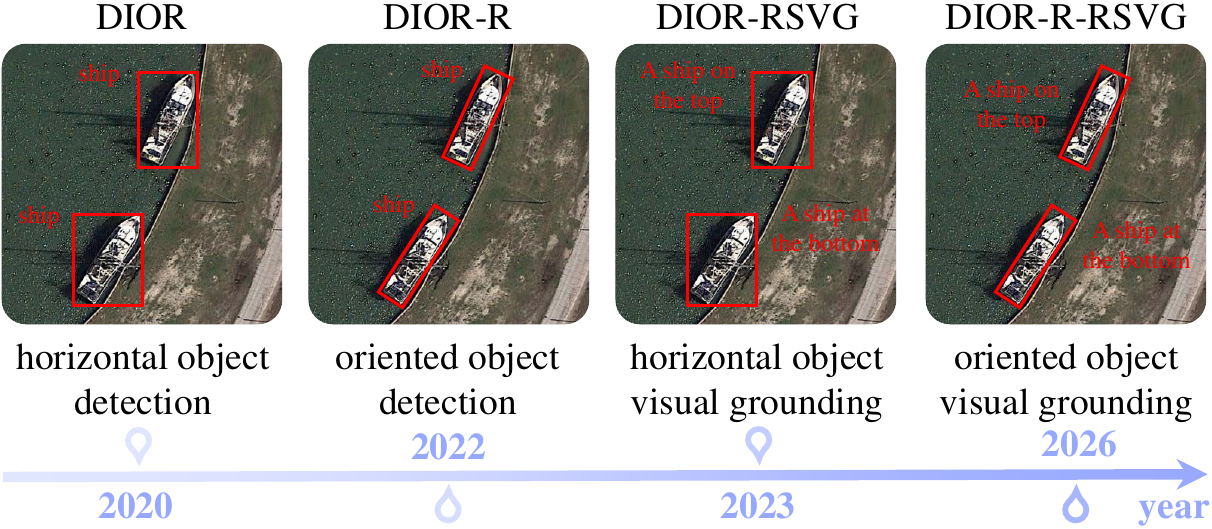}
      \caption{The evolution of datasets. DIOR~\cite{dior} is a horizontal object detection dataset. DIOR-R~\cite{diorr} is an oriented object detection dataset. DIOR-RSVG~\cite{diorrsvg} is a horizontal object visual grounding dataset. DIOR-R-RSVG is an oriented object visual grounding dataset.}
      \label{fig:diorrrsvghistory}
    \end{figure}
}

\newcommand{\otwovgvlmtrainloss}{
    \begin{figure}[!t]
      \centering
      \includegraphics[width=\linewidth]{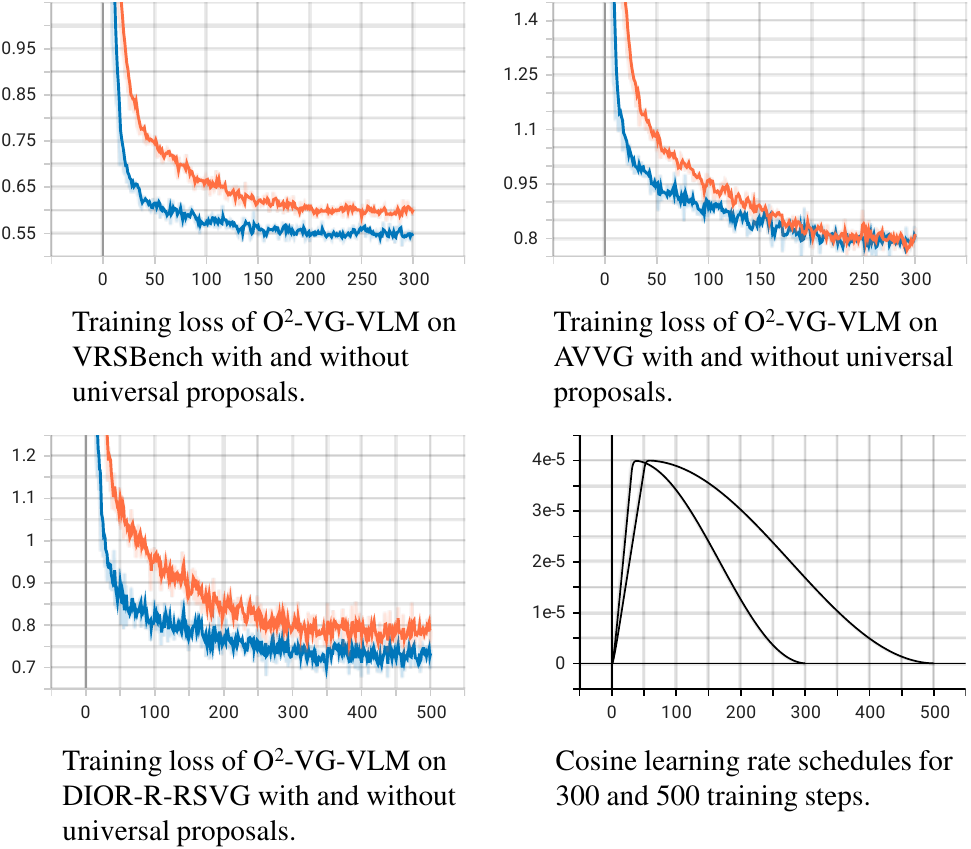}
      \caption{Training loss of O$^2$-VG-VLM on the VRSBench, AVVG, and DIOR-R-RSVG datasets with universal proposals (blue) and without universal proposals (orange). The bottom-right plot shows cosine learning rate schedules with 300 and 500 training steps.
      }
      \label{fig:otwovgvlmtrainloss}
    \end{figure}
}

\newcommand{\otwovgfamilyvisual}{
    \begin{figure*}[!t]
    \centering
    \includegraphics[width=\linewidth]{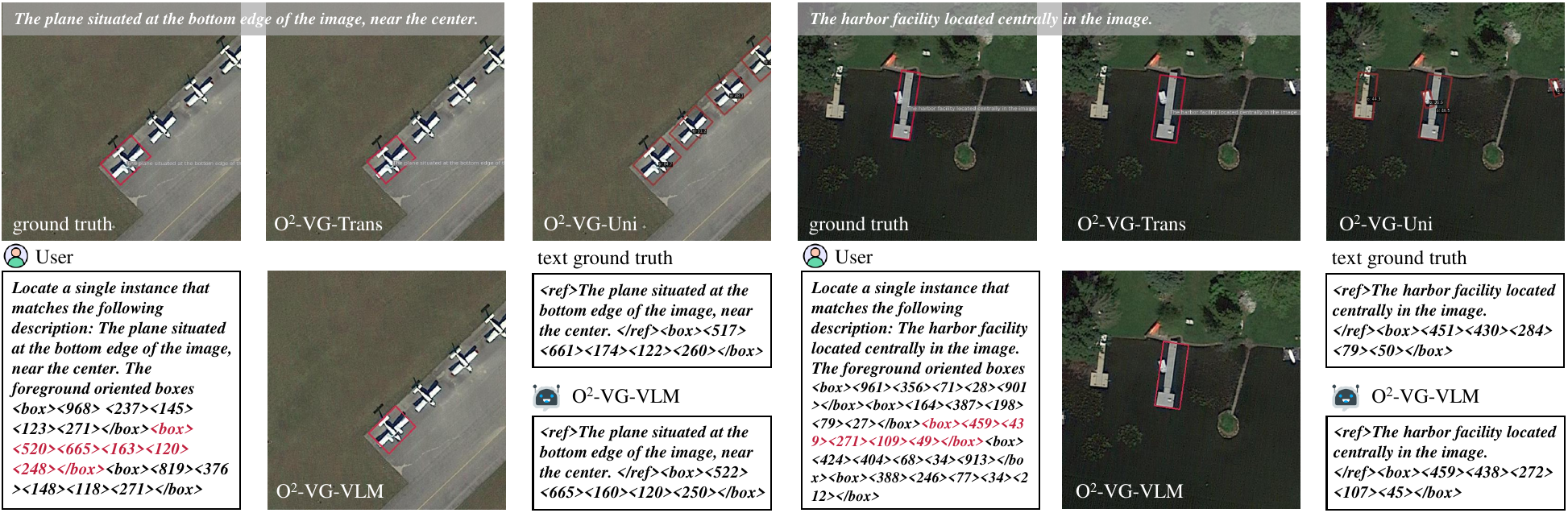}
    \caption{Visualization of the overall O$^2$-VG family paradigm for oriented object visual grounding on VRSBench. Specifically, the referring expression, ground-truth oriented box, O$^2$-VG-Trans output, universal oriented proposals produced by O$^2$-VG-Uni, input prompt, text ground truth, and O$^2$-VG-VLM output are shown.}
    \label{fig:otwovgfamilyvisual}
    \end{figure*}
}

\newcommand{\otwovgtransresultvisual}{
    \begin{figure}[!t]
      \centering
      \includegraphics[width=\linewidth]{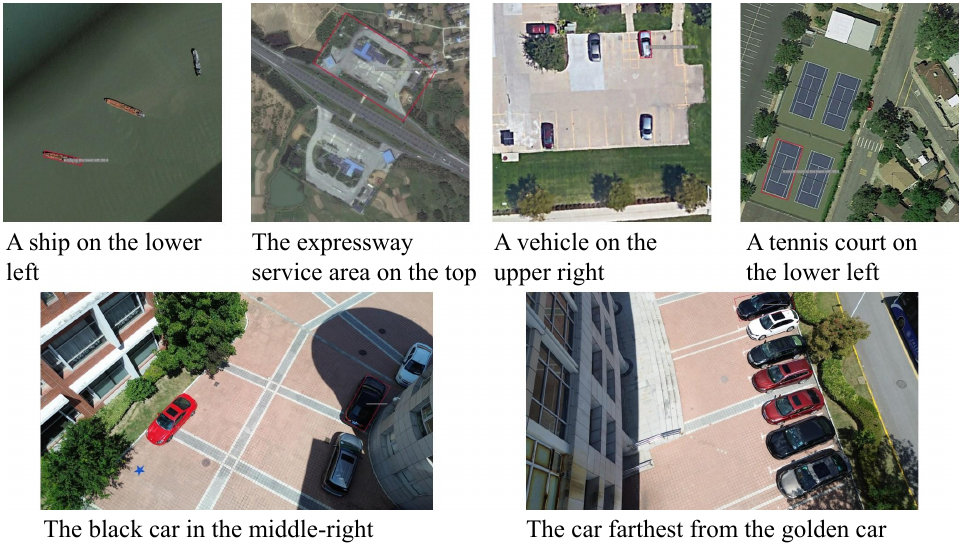}
      \caption{Top row: Visualization of O$^2$-VG-Trans results on DIOR-R-RSVG for the oriented object visual grounding task. Bottom row: Visualization of O$^2$-VG-Trans results on AVVG.}
      \label{fig:otwovgtransresultvisual}
    \end{figure}
}

\newcommand{\otwovguniresultvisual}{
    \begin{figure}[!t]
      \centering
      \includegraphics[width=\linewidth]{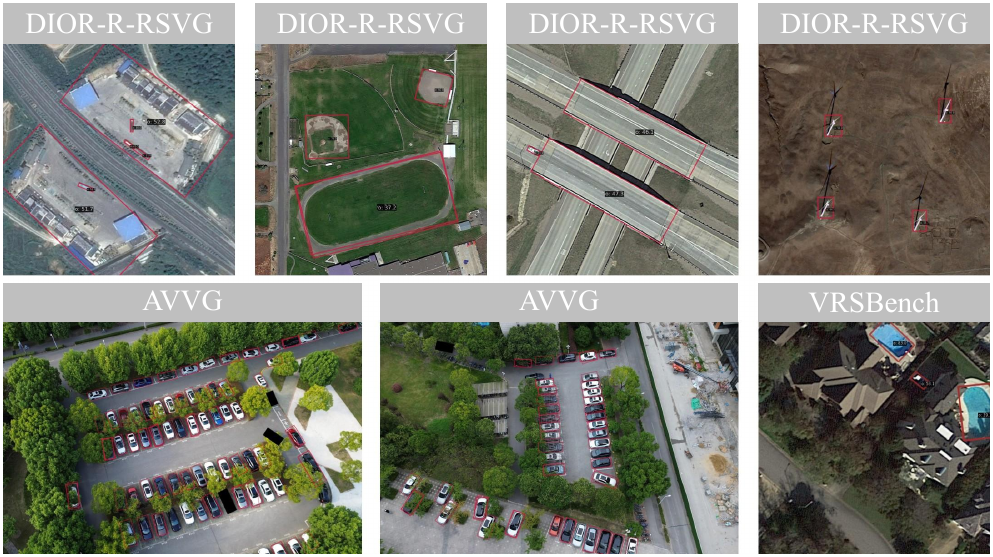}
      \caption{Visualization of the universal oriented object proposals produced by O$^2$-VG-Uni on DIOR-R-RSVG, AVVG, and VRSBench.}
      \label{fig:otwovguniresultvisual}
    \end{figure}
}

\newcommand{\otwovguniretrievalvisual}{
    \begin{figure}[!t]
      \centering
      \includegraphics[width=\linewidth]{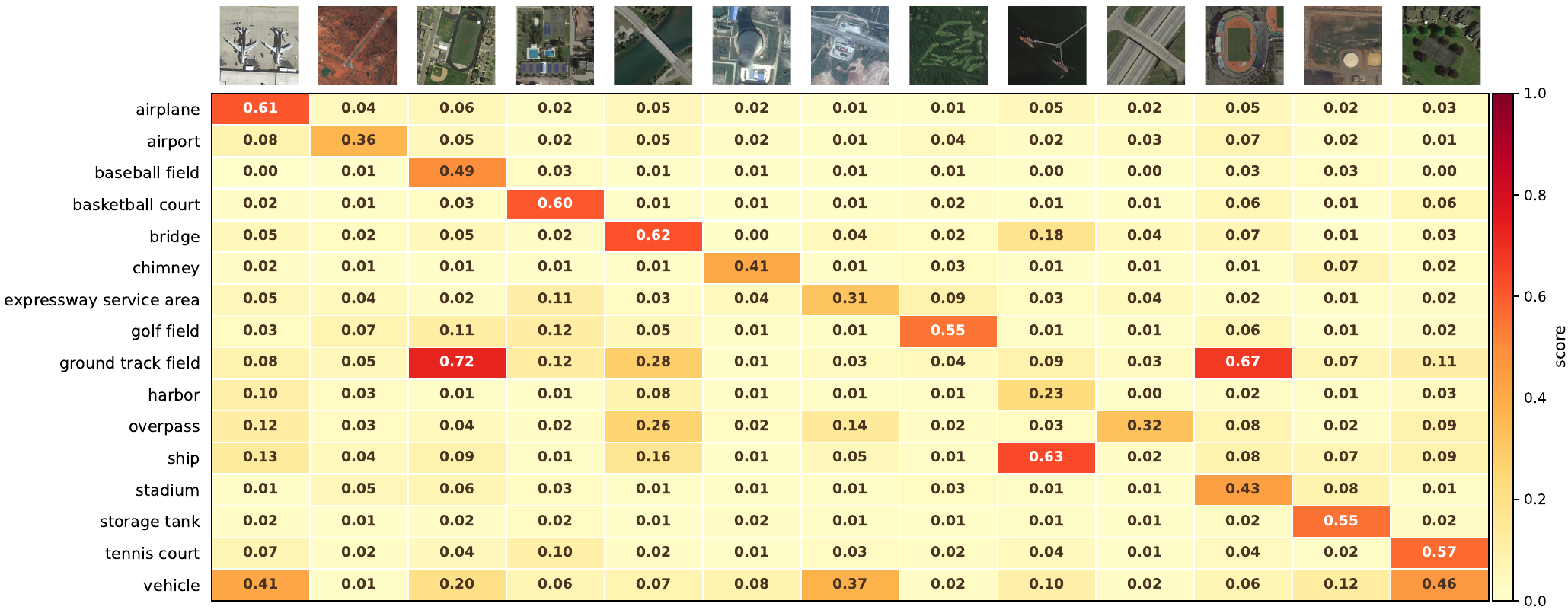}
      \caption{Visualization of O$^2$-VG-Uni object retrieval on DIOR-R-RSVG.}
      \label{fig:otwovguniretrievalvisual}
    \end{figure}
}

\newcommand{\otwovgvlmresultvisual}{
    \begin{figure}[!t]
      \centering
      \includegraphics[width=\linewidth]{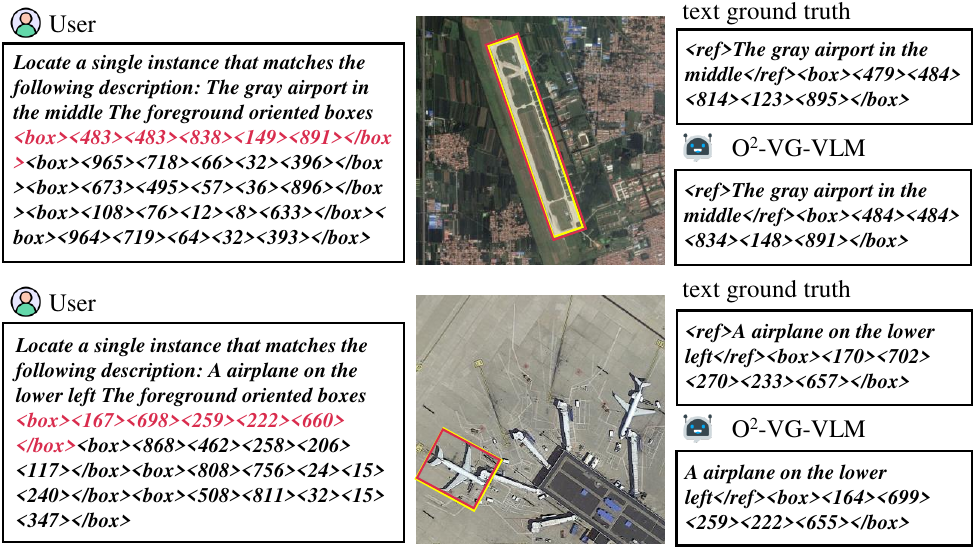}
      \caption{Visualization of the O$^2$-VG-VLM input prompt, output, and text ground truth on DIOR-R-RSVG. The yellow oriented box denotes the ground truth, while the red oriented box denotes the predicted box.}
      \label{fig:otwovgvlmresultvisual}
    \end{figure}
}

\begin{document}

\title{A Unified Framework and Dataset for Oriented Object Visual Grounding in Remote Sensing}

\author{
Zeyu Ding\textsuperscript{~\orcidlink{0009-0005-9614-7131}}, 
Yong Zhou\textsuperscript{~\orcidlink{0000-0001-6207-0299}}, 
Jiaqi Zhao\textsuperscript{~\orcidlink{0000-0002-3564-5090}},~\IEEEmembership{Member,~IEEE}, 
Wen-Liang Du\textsuperscript{~\orcidlink{0000-0002-9234-0912}}~\IEEEmembership{Member,~IEEE},\\ 
Xixi Li~\IEEEmembership{Member,~IEEE},
Hancheng Zhu\textsuperscript{~\orcidlink{0000-0002-5418-9879}},
Rui Yao\textsuperscript{~\orcidlink{0000-0003-2734-915X}},~\IEEEmembership{Member,~IEEE}, 
and Abdulmotaleb El  Saddik\textsuperscript{~\orcidlink{0000-0002-7690-8547}},~\IEEEmembership{Fellow,~IEEE}

\thanks{
Received September 2026; revised **; accepted  **. Date of publication  2026; date of current version 2026. This work was supported in part by the National Natural Science Foundation of China (Grant No. 62272461, 62676414, 62602671, 62671592) \textit{(Corresponding author: Yong Zhou.)}

Zeyu Ding, Yong Zhou, Jiaqi Zhao, Wen-Liang Du, Xixi Li, Hancheng Zhu, and Rui Yao are with the School of Computer Science and Technology/School of Artificial Intelligence, the Mine Digitization Engineering Research Center of the Ministry of Education, and Jiangsu Provincial Industrial Technology Engineering Center for Intelligent Sensing and Emergency IoT in Underground Space, China University of Mining and Technology, Xuzhou 221116, China (e-mails: dingzeyu@cumt.edu.cn, yzhou@cumt.edu.cn, jiaqizhao@cumt.edu.cn, wldu@cumt.edu.cn, xixil@cumt.edu.cn, zhuhancheng@cumt.edu.cn, ruiyao@cumt.edu.cn).}

\thanks{Abdulmotaleb El Saddik is with the School of Electrical Engineering and Computer Science, University of Ottawa, Ottawa, ON K1N 6N5, Canada (e-mail: elsaddik@uottawa.ca).}

\thanks{Digital Object Identifier }

}

\markboth{IEEE Transactions on Geoscience and Remote Sensing,~Vol.~XX, No.~XX, 2026}%
{Ding \MakeLowercase{\textit{et al.}}: Parallel Oriented Box Generation}

\maketitle

\begin{abstract}
Visual grounding in remote sensing images aims to locate objects described by referring expressions. Most existing methods predict horizontal bounding boxes, which are often inaccurate for objects with arbitrary orientations. To address this limitation, we introduce O$^2$-VG, a family of models for oriented object visual grounding with three complementary designs.
Specifically, O$^2$-VG-Trans is a cross-modality transformer for oriented object visual grounding. It establishes a strong discriminative foundation for the model family. Building upon it, O$^2$-VG-Uni predicts universal oriented proposals for possible foreground objects without specific text prompts. It also supports object retrieval through cached proposal embeddings. Using these universal oriented proposals as input prompts, O$^2$-VG-VLM is an autoregressive vision-language model. It generates oriented box token blocks in parallel through multi-token prediction.
In addition, we construct DIOR-R-RSVG, a dataset for oriented object visual grounding in remote sensing images. It provides image, expression, and oriented box triplets for training and evaluation. Together, the O$^2$-VG family provides a flexible framework that spans discriminative transformers and generative vision-language models. It achieves superior performance across multiple benchmarks. Code is available at {\textcolor[HTML]{181717}{\raisebox{-0.25\height}{\simpleicon{github}}}} \url{https://github.com/wokaikaixinxin/ai4rs} and {\textcolor[HTML]{181717}{\raisebox{-0.25\height}{\simpleicon{github}}}} \url{https://github.com/wokaikaixinxin/Eagle_o2_vg}.
\end{abstract}


\teaser

\section{Introduction}

Visual grounding, also known as referring expression comprehension~\cite{RRSECS}, aims to locate target objects in an image based on referring expressions.
It enables language-guided object localization by associating visual regions with textual descriptions.
Remote sensing images provide rich visual information of geographic scenes and support various applications~\cite{rqformer}.
Compared with natural images~\cite{coco}, remote sensing images contain objects with scale variations, dense distributions, and arbitrary orientations~\cite{diorrsvg}.
Users often need to locate target objects through descriptions of their categories, attributes, or locations.
Therefore, visual grounding in remote sensing images has become an important task for connecting language descriptions with object localization.

Most visual grounding methods return a horizontal bounding box (HBB)~\cite{10542207,geovg,effgrounddino}. This representation leads to inaccurate localization with excessive background regions for oriented objects. Oriented bounding boxes (OBBs) provide a tighter representation by explicitly modeling the object center, size, and angle~\cite{rediffdet,orientedformer,gsdet}, as illustrated in Fig.~\ref{fig:teaser}. However, oriented object visual grounding requires the model to understand both the meaning of the expression and the geometric structure of the target. It also requires data that pairs language expressions with oriented annotations.

Existing remote sensing visual grounding (RSVG) methods mainly use discriminative architectures~\cite{effgrounddino}. They match text features with image regions and then regress a box. These methods usually depend on horizontal proposals and task specific heads. Generative vision language models offer a more flexible interface~\cite{geochat,refgeo}. Both visual and text inputs are modeled under the next-token prediction paradigm. Their autoregressive decoding, however, produces coordinate tokens one at a time. This can be slow when an image contains many targets. It may also produce invalid box structures. These limitations motivate a unified study of oriented object visual grounding.

\overview

In this work, we present O$^2$-VG, a model family for oriented object visual grounding in remote sensing images, as illustrated in Fig.~\ref{fig:overview}. Specifically, we first build O$^2$-VG-Trans, a strong discriminative end-to-end cross-modality transformer, which serves as the foundation of the O$^2$-VG family. Building upon it, O$^2$-VG-Uni produces universal oriented proposals for possible foreground objects without specific text prompts and supports object retrieval with proposal embeddings. O$^2$-VG-VLM is an autoregressive vision-language model that uses these universal proposals as text prompts. It predicts oriented box tokens in parallel within each block while maintaining a causal order between blocks. A hybrid inference strategy further combines parallel decoding with token-by-token decoding when invalid blocks are generated. To facilitate oriented object visual grounding, we construct a new dataset, DIOR-R-RSVG. It contains image, expression, and oriented box triplets for training and evaluation.
We evaluate the proposed models and dataset on multiple benchmarks to validate their effectiveness for oriented object visual grounding in remote sensing images.

The contributions of this paper are summarized as follows.
\begin{itemize}
    \item We propose O$^2$-VG-Trans, a discriminative cross-modality transformer framework for oriented object visual grounding, which establishes a strong foundation for the O$^2$-VG family.
    \item We design O$^2$-VG-Uni, a universal oriented proposal framework that provides potential foreground proposals without specific text prompts and supports object retrieval through proposal embeddings.
    \item We develop O$^2$-VG-VLM, an autoregressive vision-language grounding model that generates oriented box tokens through multi-token prediction within a unified generation framework.
    \item We introduce DIOR-R-RSVG, a new oriented object visual grounding dataset containing remote sensing image, referring expression, and oriented box triplets for training and evaluation.
\end{itemize}

\section{Related Work}

\subsection{Discriminative Remote Sensing Visual Grounding}

Remote sensing visual grounding aims to localize the object referred to by an expression. Existing discriminative methods can be broadly grouped into graph-based, CNN-based, and Transformer-based approaches according to their paradigm.

\textit{Graph-based methods} construct geospatial relations among candidate regions to reduce the search space in large-scale scenes. GeoVG~\cite{geovg} represents referring expressions as relation graphs and matches them with image regions. 
\textit{CNN-based methods} adopt two-stage or one-stage designs. Two-stage methods use CNN-extracted proposals and select the region with the highest visual-text matching score~\cite{zhu2025cascaded}, so performance depends on proposal recall and incurs matching cost. One-stage methods inject language features into CNN maps and directly regress horizontal boxes from dense locations~\cite{openrsd}. 
\textit{Transformer-based methods} use self-attention and cross-attention to capture long-range dependencies between image regions and language tokens. By jointly modeling global context and fine-grained visual-text interactions, they improve grounding robustness in cluttered scenes and under complex referring expressions~\cite{10542207,effgrounddino}.

Existing discriminative remote sensing visual grounding methods focus on horizontal boxes, which cannot accurately describe objects with arbitrary orientations. Unlike existing methods, our approach performs oriented object visual grounding, enabling more accurate localization of arbitrarily rotated targets.

\subsection{Generative VLM for Remote Sensing Visual Grounding}

Generative autoregressive vision-language models unify visual grounding and textual generation within a single next-token prediction framework. 

Among existing generative RSVG methods with oriented-box outputs, GeoChat~\cite{geochat} follows the LLaVA-v1.5~\cite{llava} architecture, coupling a CLIP-ViT~\cite{clip} visual encoder with a two-layer MLP adaptor and a Vicuna language decoder. It pioneers instruction-following grounding by treating object localization and orientation prediction as text generation, while also supporting diverse remote sensing vision-language tasks. GeoGround~\cite{refgeo} similarly adopts a CLIP-ViT visual encoder, a two-layer MLP connector, and a Vicuna decoder. It provides a unified framework for HBB, OBB, and mask grounding, using a Text-Mask representation to support flexible output formats within a single VLM.

Unlike these methods, our vlm adopts multi-token prediction to generate the complete token block of each oriented box in parallel while preserving causal dependencies across blocks.

\section{Method}
\label{sec:method}

In this section, we present the family of models for oriented object visual grounding in remote sensing imagery. O$^2$-VG-Trans (Sec.~\ref{sec:o2vgtrans}) serves as the foundational framework, a cross-modality transformer. Building upon it, O$^2$-VG-Uni (Sec.~\ref{sec:o2vguni}) removes the reliance on text prompts, evolving into a universal oriented proposal generator that supports fast retrieval. O$^2$-VG-VLM (Sec.~\ref{sec:o2vgvlm}) leverages the oriented proposals as part of the input prompts to an autoregressive vision-language model with multi-token coordinate prediction.

\otwovgtrans

\subsection{O$^2$-VG-Trans: An Oriented Object Visual Grounding Transformer}
\label{sec:o2vgtrans}
In this section, we present O$^2$-VG-Trans, an oriented object visual grounding transformer for remote sensing images, as illustrated in Fig.~\ref{fig:o2vgtrans}.

\textit{Image Backbone.}
Given an RGB input image $\bm{I}$ $\in \mathbb{R}^{3 \times H \times W}$, we utilize ResNet~\cite{resnet} as the image backbone to extract multi-scale visual features. A convolutional Channel Mapper then projects these features to a unified channel dimension of 256, producing four visual feature maps $\{\bm{P}_i\}_{i=3}^6$, where $\bm{P}_i \in \mathbb{R}^{256 \times H_i \times W_i}$ at strides of $\{8, 16, 32, 64\}$, respectively.

\textit{Text Backbone.}
Given an input referring expression of $N_t$ tokens, we employ a frozen pretrained BERT encoder~\cite{bert} with 12 layers and a hidden dimension of 768 to extract textual features. 
Sub-sentence level attention is adopted with a custom mask to restrict self-attention within each sub-sentence. The BERT outputs are then projected to 256 channels via a linear layer, yielding  textual embedding $\bm{T} \in \mathbb{R}^{N_t \times 256}$.

\textit{Cross-modality Encoder.}
The cross-modality encoder consists of six stacked blocks that enable visual and textual interaction.
Specifically, the visual feature maps $\{\bm{P}_i\}_{i=3}^6$ are flattened and concatenated to form the visual sequence $\bm{V} \in \mathbb{R}^{N_v \times 256}$. 
At each encoder layer, a bidirectional attention block first exchanges information across modalities:
\begin{equation}
(\bm{F}_V, \bm{F}_T) = \texttt{BiAttnBlock}(\bm{V}, \bm{T}).
\end{equation}
Then, the updated features are independently refined via modality-specific attention layers:
\begin{equation}
\bm{X}_V = \texttt{DefAttn}(\bm{F}_V), \quad \bm{X}_T = \texttt{SAttn}(\bm{F}_T),
\end{equation}
where $\texttt{DefAttn}(\cdot)$ and $\texttt{SAttn}(\cdot)$ correspond to deformable attention layer and self-attention layer, respectively.

\textit{Language-Guided Query Selection.}
Given the visual embeddings $\bm{X}_V \in \mathbb{R}^{N_v \times 256}$ and the text embeddings $\bm{X}_T \in \mathbb{R}^{N_t \times 256}$, we first compute the region-text matching score matrix $\bm{M} \in \mathbb{R}^{N_v \times N_t}$ via cosine similarity:
\begin{equation} \label{cos_similarity}
\bm{M} = \frac{\bm{X}_V \cdot \bm{X}_T^\top}{\|\bm{X}_V\|_2 \|\bm{X}_T\|_2}.
\end{equation}
Then we execute the $\max(\cdot)$ operation on $\bm{M}$ along the $-1$ dimension. The top-$K$ oriented reference boxes $\bm{B} \in \mathbb{R}^{K \times 5}$, along with learnable queries $\bm{Q}\in \mathbb{R}^{K\times256}$, are initialized for the decoder, where $K$ is set to 900 by default.

\textit{Cross-Modality Decoder.}
The decoder consists of six stacked blocks that inject textual and visual features into queries through different layers:
\begin{equation}
\bm{Q} = \texttt{FFN} \left( \texttt{DefAttn} \left( \texttt{CAttn} \left( \texttt{SAttn}(\bm{Q}), \bm{X}_T \right), \bm{X}_V, \bm{B} \right) \right),
\end{equation}
where $\texttt{CAttn}(\cdot)$ and $\texttt{FFN}(\cdot)$ denote a text cross-attention layer and a feed-forward network, respectively. In the deformable attention layer, sampling points are rotated around the centers of oriented reference proposals $\bm{B}$.

\textit{Head and Loss.}
The prediction head consists of a contrastive classification branch and a regression branch. For classification, cosine similarity scores between the updated queries $\bm{Q}$ and text embeddings $\bm{X}_T$ are computed as $\bm{Q}\cdot\bm{X}_T^\top/(\|\bm{Q}\|_2\|\bm{X}_T\|_2)$ following Eq.~\ref{cos_similarity}. For regression, a Multi-Layer Perceptron predicts oriented boxes. We adopt one-to-one Hungarian matching following \cite{o2rtdetr}. The total loss is formulated as:
\begin{equation}
\mathcal{L} = \lambda_{\text{cls}} \mathcal{L}_{\text{cls}} + \lambda_{\text{L1}} \mathcal{L}_{\text{L1}} + \lambda_{\text{iou}} \mathcal{L}_{\text{iou}},
\end{equation}
where $\mathcal{L}_{\text{cls}}$ is Focal Loss~\cite{focal_loss} for contrastive classification, $\mathcal{L}_{\text{L1}}$ is L1 loss, and $\mathcal{L}_{\text{iou}}$ denotes KLD loss~\cite{kld}. The loss weights are set to $\lambda_{\text{cls}} = 1.0$, $\lambda_{\text{L1}} = 5.0$, and $\lambda_{\text{iou}} = 2.0$, respectively.

\otwovguni

\subsection{O$^2$-VG-Uni: A Universal Oriented Proposal Generator}
\label{sec:o2vguni}

Building upon the foundation of O$^2$-VG-Trans, we introduce O$^2$-VG-Uni, a universal oriented proposal generator without the need for special text prompts, as illustrated in Fig.~\ref{fig:o2vguni}. This design enables two capabilities: (1) extracting universal oriented objects without specific text information, and (2) supporting fast object retrieval through cached proposal embeddings.

\textit{Two-Stage Training Recipe.}
In the first stage, only the pretrained text encoder is frozen while the entire visual branch is trained, aligning vision and language through contrastive learning. In the second stage, we freeze the entire visual branch with weights inherited from the first stage, and replace the text encoder with the learnable objectness prompt $\bm{E}_{obj}$ to extract objects of arbitrary categories.

\textit{Image Backbone.}
Given an input RGB image $\bm{I}\in\mathbb{R}^{3 \times H \times W}$, we adopt ResNetvd~\cite{resnetd} as the image backbone to extract three feature maps. A convolutional Channel Mapper then projects these features into a uniform 256-dimensional space, producing $\{\bm{P}_i\}_{i=3}^5$, where $\bm{P}_i \in \mathbb{R}^{256 \times H_i \times W_i}$ corresponds to strides of $\{8, 16, 32\}$, respectively. This design employs one fewer feature layer than O$^2$-VG-Trans.

\textit{Text Encoder.}
Given a set of $N_c$ category names, we employ a frozen pretrained RemoteCLIP~\cite{remoteclip} text encoder to produce category-level semantic embeddings. The encoder processes each prompt as an independent input sequence and outputs a text embedding matrix $\bm{C} \in \mathbb{R}^{N_c \times 512}$, where each row corresponds to one category.

\textit{Hybrid Encoder.} Following the hybrid encoder paradigm from~\cite{o2rtdetr}, the encoder decouples intra-scale interactions and cross-scale fusion. A self-attention block is applied to the top-level feature map $\bm{P}_5$, while a CNN module enables efficient fusion of multi-scale features, producing enhanced visual features $\bm{V} \in \mathbb{R}^{N_v \times 256}$. Different from the cross-modality encoder in O$^2$-VG-Trans, it operates only on image features.

\textit{Learnable Objectness Prompt.}
To obtain universal oriented proposals, we introduce a learnable objectness prompt $\bm{E}_{obj} \in \mathbb{R}^{1 \times 512}$. This embedding $\bm{E}_{obj}$ represents only objectness, indicating whether a region contains a foreground object.

\textit{Query Selection.} To unify dimensions, the visual features $\bm{V}$ are first projected by a weight matrix to obtain visual embeddings \(\bm{X}_V \in \mathbb{R}^{N_v \times 512}\). We then compute a similarity matrix \(\bm{M}\) between \(\bm{X}_V\) and the guiding embeddings as:
\begin{equation}
\bm{M} =
\begin{cases}
\dfrac{\bm{X}_V \cdot \bm{C}^{\top}}{\|\bm{X}_V\|_2  \|\bm{C}\|_2^{\top}} \in \mathbb{R}^{N_v \times N_c}, & \text{stage 1}, \\[8pt]
\dfrac{\bm{X}_V \cdot \bm{E}^{\top}}{\|\bm{X}_V\|_2  \|\bm{E}\|_2^{\top}} \in \mathbb{R}^{N_v \times 1}, & \text{stage 2}.
\end{cases}
\end{equation}
By taking the maximum value along the text dimension of $\bm{M}$, we select the top-$K$ candidates to form the oriented reference boxes $\bm{B} \in \mathbb{R}^{K \times 5}$ and query embeddings $\bm{Q}\in \mathbb{R}^{K\times256}$, which are then fed into the decoder, where $K$ is set to 300 by default.

\textit{Image decoder.} Following the decoder paradigm from~\cite{o2rtdetr}, the image decoder consists of multiple stacked decoder layers, each containing self-attention, cross-attention, and feed-forward networks. Different from the cross-modality decoder in O$^2$-VG-Trans, it operates only on image modality. 

\textit{Head and Loss.} The prediction head consists of a contrastive classification branch and a regression branch. For classification, the updated queries $\bm{Q}$ are first projected using a weight matrix to obtain proposal embeddings $\bm{E}_Q \in \mathbb{R}^{K \times 512}$. In training stage 1, cosine similarity scores between the proposal embeddings $\bm{E}_Q$ and text embeddings $\bm{C}$ are computed following Eq.~\ref{cos_similarity}, while in training stage 2, cosine similarity scores between the proposal embeddings $\bm{E}_Q$ and objectness embeddings $\bm{E}_{obj}$ are computed similarly. For regression, a Multi-Layer Perceptron predicts oriented boxes. The loss settings are similar to those of O$^2$-VG-Trans.

\textit{Object Retrieval.} Unlike other class-agnostic proposal networks~\cite{faster_rcnn}, our proposal embeddings retain semantic information and can therefore be used for object retrieval. The oriented proposal embeddings $\bm{E}_Q$ can be cached, while the text embeddings generated by RemoteCLIP are used as retrieval queries. We compute the cosine similarity between the cached proposal embeddings and the text embeddings to realize object retrieval.

\otwovgvlm

\subsection{O$^2$-VG-VLM: Vision-Language Grounding with Multi-Token Prediction}
\label{sec:o2vgvlm}

O$^2$-VG-VLM is built upon an autoregressive vision-language model~\cite{locateanything} pre-trained on large-scale image-text corpora. The architecture consists of three main components: a Moon-ViT \raisebox{-0.15\height}{\includegraphics[height=1em]{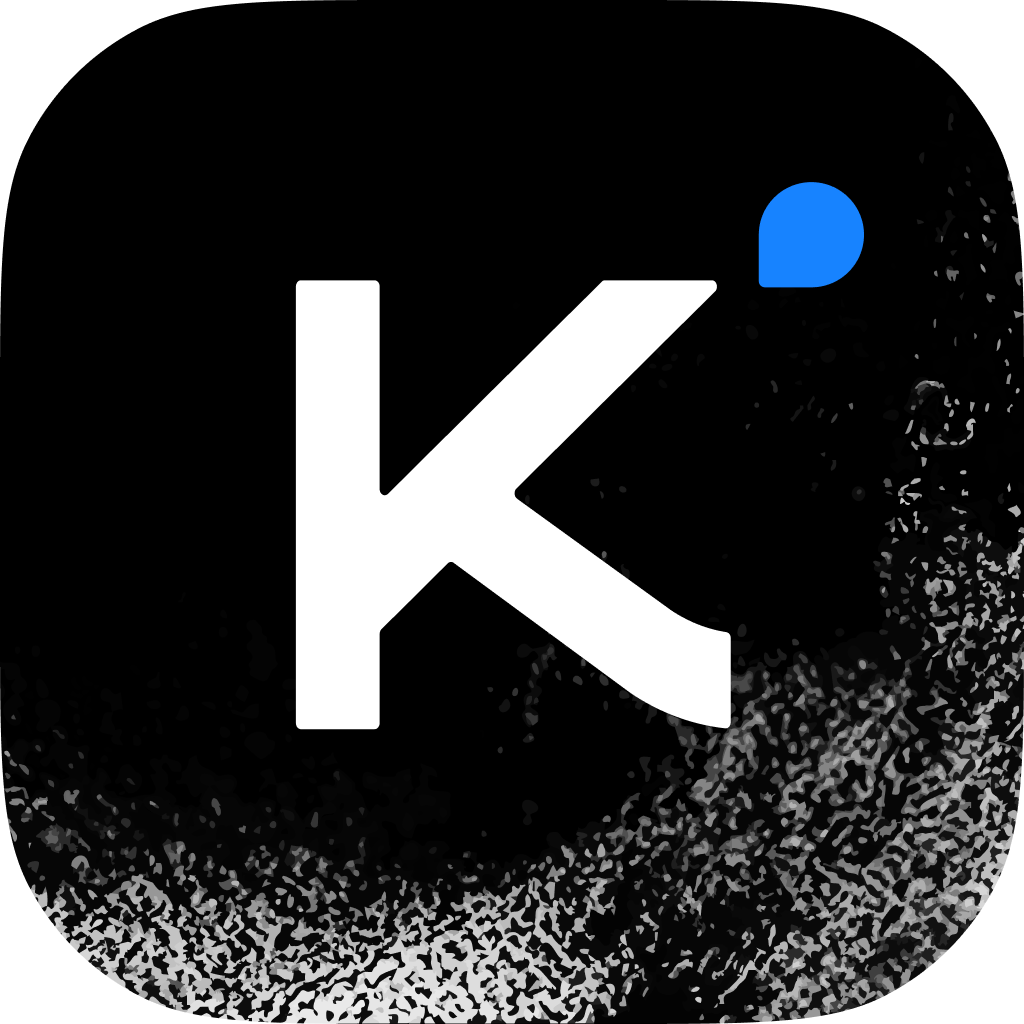}} vision encoder~\cite{kimivl}, an MLP projector, and a Qwen2.5 \raisebox{-0.15\height}{\includegraphics[height=1em]{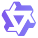}} language decoder~\cite{qwen3vl}, as shown in Fig.~\ref{fig:o2vgvlm}.

\textit{Vision Encoder.}
The Moon-ViT \raisebox{-0.15\height}{\includegraphics[height=1em]{kimi-icon-rounded-corner.png}} vision encoder splits the input image $\bm{I}\in\mathbb{R}^{3\times H\times W}$ into \(14 \times 14\) patches and encodes them with a 27-layer Transformer with 16 attention heads and hidden dimension $1152$. The resulting visual features are then merged with a \(2 \times 2\) kernel to reduce the visual token length, yielding $\bm{Z}_I\in\mathbb{R}^{N_v\times1152}$ before being fed into the projector.

\textit{Projector.}
The visual tokens produced by MoonViT are first adapted to the language decoder through a lightweight MLP projector $f_{\mathrm{proj}}$. This projector consists of two MLP connector layers and maps MoonViT features from the 1152-dimensional vision space to the 2048-dimensional hidden space of Qwen2.5, giving $\bm{Z}_V=f_{\mathrm{proj}}(\bm{Z}_I)\in\mathbb{R}^{N_v\times2048}$.

\textit{Language Decoder.}
The Qwen2.5 language decoder \raisebox{-0.15\height}{\includegraphics[height=1em]{qwen-color.pdf}} serves as the generative core of the model. It has a hidden size of 2048, 36 transformer layers, and 16 attention heads. It receives the projected visual embeddings $\bm{Z}_V$ together with the text embeddings $\bm{Z}_T$.

\textit{Convert Oriented Boxes to Tokens.}
To align continuous box coordinates with the discrete token-based output space, each component of $(cx, cy, w, h, \theta)$ is discretized into 1000 bins within $[0, 1000]$. Specifically, $(cx, w)$ and $(cy, h)$ are scaled by the image width $W$ and height $H$, respectively, while the angle $\theta$ is scaled by its predefined range. For instance, the discrete center coordinate is computed as $(\mathrm{round}(1000 \times cx / W), \mathrm{round}(1000 \times cy / H))$.

\textit{Conversation Template.}
To enable multi-modal instruction tuning, we construct a structured conversation template. An example is illustrated below.

\begin{tcolorbox}[
    colback=gray!5,
    colframe=gray!80,
    boxrule=0.8pt,
    arc=4pt,
    left=2pt,
    right=2pt,
    top=2pt,
    bottom=2pt,
    fontupper=\small
]
\textbf{human}:
Locate a single instance that matches the following description:
The plane is on the left side of the image. The foreground oriented boxes: \texttt{<box><798><588><467><378><178></box><box> <273><306><468><379><183></box>}

\textbf{gpt}:
\texttt{<ref>}The plane is on the left side of the image.
\texttt{</ref><box><272><308><468><380><183></box>}

\textbf{image}:
12345.png
\end{tcolorbox}

Specifically, we extend the vocabulary with special tokens, including \texttt{<box>}, \texttt{</box>}, \texttt{<ref>}, and \texttt{</ref>}, to delineate bounding box coordinates and textual expressions, respectively. The input foreground oriented boxes are provided by O$^2$-VG-Uni (Sec.~\ref{sec:o2vguni}).

\textit{Next-Token Prediction.}
Let the answer sequence generated by the next-token prediction (NTP) be denoted as $x_{1:N_{\text{ntp}}}$. The NTP objective maximizes the conditional likelihood of each token given all preceding tokens:
\begin{equation}
    \mathcal{L}_{\text{NTP}}=-\sum_{n=1}^{N_{\text{ntp}}} \log P(x_{n+1}\mid x_{1:n},\bm{Z}_V,\bm{Z}_T),
\end{equation}
where the sum ranges over every answer position.

\textit{Causal Attention for NTP.}
For next-token prediction, O$^2$-VG-VLM employs a causal attention mechanism, which ensures that each token can only attend to prior tokens. By masking future positions, the attention preserves the left-to-right order of the generated sequence, making it naturally suited for language decoding tasks.

\textit{Multi-Token Prediction.}
Each block has a fixed length of seven tokens. A Referring Expression Block encodes the referring expression. When the expression length exceeds the block length, it is split into consecutive blocks. An Oriented Box Block comprises five tokens representing the oriented bounding box $(cx,cy,w,h,\theta)$ and two special tokens, e.g., \texttt{<box>} and \texttt{</box>}. An End Block signals the termination of generation. Any unused positions within a block are padded with a \texttt{<null>} token.
The MTP objective predicts a future block of 7 tokens in parallel:
\begin{equation}
\mathcal{L}_{\mathrm{MTP}}
=
-\sum_{n \in \mathcal{I}_{\mathrm{mtp}}}
\sum_{m=1}^{7}
\log
P\left(
x_{n+m}
\mid
x_{1:n},
\mathbf{Z}_{V},
\mathbf{Z}_{T}
\right),
\end{equation}
where $\mathcal{I}_{\mathrm{mtp}}$ contains the token positions that serve as the starting context for MTP blocks.

\diorrrsvgexpression
\diorrrsvgobb

\textit{Bidirectional Intra-Block Attention.}
Within each MTP block, we employ bidirectional attention, allowing all seven positions to attend to one another. This design enables the model to jointly capture the structural dependencies among tokens within a block, such as the dependencies between the coordinate tokens and boundary tokens of an oriented box, and to predict the entire block simultaneously.

\textit{Causal Flow Across Blocks.}
Across MTP blocks, attention follows a causal order. The current block can attend to the visual-textual context and all previously committed tokens, but not to tokens in subsequent blocks. This block-causal design preserves the sequential dependency between different referring-expression, box, and end blocks, while retaining parallel prediction within each block.

\textit{Inference Modes.}
O$^2$-VG-VLM supports three inference modes. \emph{Slow Mode} uses standard NTP to generate tokens autoregressively, providing the most reliable decoding. \emph{Fast Mode} uses MTP to generate seven-token blocks in parallel for maximum throughput. \emph{Hybrid Mode} performs MTP by default and validates each generated block. When an invalid box structure is detected, the current block is discarded and the model switches to NTP to decode the problematic box token by token. Specifically, an invalid coordinate is identified when the top-1 coordinate token's probability is below 0.9 and the span of the valid top-5 coordinate tokens exceeds 60 within the [0, 1000] normalized space. After the box is completed, decoding resumes in MTP mode. This design combines the efficiency of parallel block decoding with the robustness of autoregressive generation.

\section{DIOR-R-RSVG dataset}
\label{sec:diorrrsvg}

We designed DIOR-R-RSVG, a new dataset containing $<\text{image}, \text{expression}, \text{obb}>$ triplets for oriented object visual grounding in remote sensing imagery.

\subsection{Data Sources and Foundation}

\textit{Images.}
Images are sourced from the DIOR-RSVG dataset~\cite{diorrsvg}, which is built for the horizontal box visual grounding task in remote sensing. The dataset contains 17,402 RGB images.

\textit{Referring expressions.}
Referring expressions are also sourced from DIOR-RSVG~\cite{diorrsvg}. The dataset comprises 38,320 text expressions, where each expression corresponds to a single object instance.

\textit{Oriented Bounding Boxes.}
Oriented box annotations are sourced from the DIOR-R dataset~\cite{diorr}, which is built for the oriented object detection task in remote sensing. It contains 23,463 images and 192,512 annotated object instances.

\subsection{Construction Pipline}

Images in DIOR~\cite{dior}, DIOR-RSVG~\cite{diorrsvg}, and DIOR-R~\cite{diorr} come from the same domain. This allows each object instance to naturally correspond across horizontal boxes, oriented boxes, and referring expressions.  We leverage this consistency to connect $< \text{image}, \text{expression}, \text{obb} >$ triplets smoothly. The procedure involves the following steps:

\begin{enumerate}
    \item \textit{Data Sampling:} We extract $< \text{image},$ $ \text{expression}, \text{hbb}, $ $\text{category} >$ tuples from DIOR-RSVG~\cite{diorrsvg} and $< \text{image}, $ $ \text{obb}, $ $ \text{category} >$ tuples from DIOR-R~\cite{dior}.
    \item \textit{Automated Matching:} For each instance in DIOR-RSVG~\cite{diorrsvg}, we calculate the IoU between its HBB and the minimum rectangle of OBBs in the same category and image. If the maximum IoU exceeds a predefined threshold, we select the corresponding OBB as the match.
    \item \textit{Human Verification and Annotation:} We manually check the automatic matching results. All unmatched instances and failure cases are then reviewed and re-annotated to guarantee data quality.
\end{enumerate}

\DIORRRSVGStatistics

\subsection{Dataset Statistics and Analysis}

We conduct a statistical analysis of the proposed DIOR-R-RSVG dataset.

\textit{Images.}
The proposed dataset contains 17,402 RGB images of size 800 $\times$ 800 pixels, with spatial resolutions ranging from 0.5 to 30 meters per pixel, as shown in Table~\ref{Tab:DIORRRSVGStatistics}. 

\textit{Referring expressions.}
The proposed dataset consists of 38,320 expressions, each corresponding to a single object instance. It is divided into 26,991 (70\%) training samples, 3,829 (10\%) validation samples, and 7,500 (20\%) testing samples, as shown in Table~\ref{Tab:DIORRRSVGStatistics}. Statistics of expression length and the word cloud visualization for the DIOR-R-RSVG dataset are shown in Fig.~\ref{fig:diorrrsvgexpression}. The average expression length is 7.5 words, and the majority of expressions contain between 3 and 18 words.

\textit{Oriented Bounding Boxes.}
Statistics of oriented object size, area, angle, and categories in the DIOR-R-RSVG dataset are shown in Fig.~\ref{fig:diorrrsvgobb}. The oriented objects correspond to the expressions, resulting in a total of 38,320 oriented boxes. Each oriented box is defined by four vertices. The objects exhibit large aspect ratios, significant scale variations, and arbitrary orientations. A total of 20 categories are included in the dataset, among which \textit{Vehicle} is the most frequent category and \textit{Harbor} is the least frequent.

\textit{Comparison with Other Datasets.}
Our DIOR-R-RSVG dataset is built upon DIOR~\cite{dior}, DIOR-R~\cite{diorr}, and DIOR-RSVG~\cite{diorrsvg}. Their evolution is shown in Fig.~\ref{fig:diorrrsvghistory}. DIOR-R-RSVG supports oriented bounding boxes and referring expressions. Furthermore, we compare DIOR-R-RSVG with other visual grounding datasets, including VRSBench and AVVG, as shown in Table~\ref{Tab:DIORRRSVGStatistics}. DIOR-R-RSVG contains a larger number of images and expressions, as well as larger image sizes, than these existing datasets.

\section{Experiments}

\subsection{Datasets}

We conduct extensive experiments on three datasets: VRSBench~\cite{vrsbench}, AVVG~\cite{geoground}, and DIOR-R-RSVG.

VRSBench~\cite{vrsbench} is a versatile vision-language benchmark for remote sensing image understanding, comprising 29,614 images with a fixed resolution of $512\times512$ for image captioning, visual grounding, and visual question answering. In our experiments, we only utilize its oriented object visual grounding annotations, which contain 36,313 expressions in the \texttt{train} set and 16,159 expressions in the \texttt{test} set.

\diorrrsvghistory

Aerial Vehicle Visual Grounding (AVVG) is a dataset introduced in GeoGround~\cite{geoground} for visual grounding of objects captured by unmanned aerial vehicles. It contains oriented bounding box annotations and covers diverse flight altitudes and scenarios. The dataset consists of 624 images with an original resolution of $4000\times2250$, containing 26,482 expressions in the \texttt{train} set and 6,139 expressions in the \texttt{test} set. In our experiments, all images are resized to $1204\times576$.

DIOR-R-RSVG is our proposed dataset for oriented object visual grounding, which is introduced in detail in Sec.~\ref{sec:diorrrsvg}. We use both the \texttt{train} and \texttt{val} sets for training, while the \texttt{test} set is used for evaluation.

\OtwoVGTransResult
\OtwoVGUniRecall

\subsection{Evaluation Metrics.}

%For each test sample, we rank the predicted oriented boxes by their prediction scores and select the highest-scoring box as the top-1 prediction. We use rotated IoU (RIoU) to measure the overlap between the predicted and ground-truth oriented boxes.

\textit{Accuracy.}
Given an IoU threshold $\tau$, the localization accuracy is defined as the proportion of samples whose top-1 prediction has an rotated IoU no lower than $\tau$:
$\mathrm{Pr}@\tau=\frac{1}{N_{\text{eval}}}\sum_{i=1}^{N_{\text{eval}}}\mathbb{I}\left[\mathrm{RIoU}_{i} \geq \tau\right]$,
where $N_{\text{eval}}$ is the number of evaluated samples and $\mathrm{RIoU}_{i}$ is the rotated IoU of the top-1 prediction for the $i$-th sample.

\textit{Mean IoU.}
Mean IoU is defined as the average of the rotated IoU values across all evaluated samples, i.e., $\mathrm{meanIoU}=\frac{1}{N_{\mathrm{eval}}}\sum_{i=1}^{N_{\mathrm{eval}}}\mathrm{RIoU}_{i}$.

\textit{Cumulative IoU.} Cumulative IoU sums the intersection and union areas over all evaluated samples before computing their ratio, defined as $\mathrm{cumIoU}={\sum_{i=1}^{N_{\text{eval}}}\mathrm{Area}(\hat{\mathcal{B}}_{i}\cap\mathcal{B}_{i})} / {\sum_{i=1}^{N_{\text{eval}}}\mathrm{Area}(\hat{\mathcal{B}}_{i}\cup\mathcal{B}_{i})}$, where $\hat{\mathcal{B}}_{i}$ and $\mathcal{B}_{i}$ denote the top-1 predicted and ground-truth oriented boxes, respectively.

\textit{Recall.}
At rotated IoU threshold $\tau$, recall is the proportion of ground-truth boxes that are successfully matched by a prediction with rotated IoU at least $\tau$.

\subsection{Implementation Details}

\textit{O$^2$-VG-Trans.}
The experiments for O$^2$-VG-Trans are conducted on two \textcolor[HTML]{76B900}{\raisebox{-0.15\height}{\simpleicon{nvidia}}} RTX 2080ti with a batch of 8 (4 images per GPU). Models are constructed based on \href{https://github.com/wokaikaixinxin/ai4rs}{AI4RS} with Pytorch \textcolor[HTML]{EE4C2C}{\raisebox{-0.15\height}{\simpleicon{pytorch}}}. 
We optimize models with the AdamW optimizer.
On VRSBench, AVVG, and DIOR-R-RSVG, we train the model for 12 epochs with an initial learning rate of $1\times10^{-4}$ and a weight decay of $1\times10^{-4}$, decaying the learning rate at the 11th epoch.

\textit{O$^2$-VG-Uni.}
We adopt the same GPU settings, batch size, loss weights, and optimizer as O$^2$-VG-Trans. In the first training stage, we train on VRSBench and DIOR-R-RSVG for 24 epochs, and on AVVG for 72 epochs. In the second training stage, we train on all three datasets for 12 epochs.

\textit{O$^2$-VG-VLM.}
The experiments for O$^2$-VG-VLM are conducted on a single \textcolor[HTML]{76B900}{\raisebox{-0.15\height}{\simpleicon{nvidia}}} RTX Pro 6000.
We initialize the model from LocateAnything-3B~\cite{locateanything} and implement the training
pipeline using PyTorch. The entire model is fine-tuned with BF16 mixed
precision. The per-device batch size is set to 1 with 16 gradient
accumulation steps. We use AdamW
with a learning rate of $4\times10^{-5}$ and a weight decay of 0.01,
together with a cosine learning rate scheduler and a 10\% warm-up ratio.
Gradient checkpointing is enabled, and the maximum sequence length is
set to 8,192 tokens. We train the model on VRSBench and AVVG for 300
optimization steps, and on DIOR-R-RSVG for 500 optimization steps,
corresponding to approximately two epochs for all datasets.

\subsection{O$^2$-VG-Trans Results for Visual Grounding}

Since no small-parameter method has been specifically designed for the oriented object visual grounding task, we adapt three horizontal object visual grounding methods by incorporating angle prediction.

GiT~\cite{git} is a vision transformer that unifies diverse vision tasks into an autoregressive framework built on the SAM ViT. Rotated GiT-B further introduces an angle token to support oriented object prediction.
Grounding DINO~\cite{groundingdino} is a grounded pre-training Transformer. Rotated GDINO further appends an angle prediction branch using CSL~\cite{csl}.
Efficient Grounding DINO~\cite{effgrounddino} improves Grounding DINO for horizontal visual grounding in remote sensing images. Rotated EGDINO further appends an angle prediction branch using CSL~\cite{csl}.

Table~\ref{Tab:OtwoVGTransResult} compares O$^2$-VG-Trans with other methods on VRSBench, AVVG, and DIOR-R-RSVG. O$^2$-VG-Trans consistently outperforms Rotated GiT, Rotated GDINO, and Rotated EGDINO across all three datasets and evaluation metrics. On VRSBench, it achieves 67.71\% Pr@0.5, 55.01\% meanIoU, and 61.14\% cumIoU. On AVVG, O$^2$-VG-Trans obtains 18.00\% Pr@0.5, 14.59\% meanIoU, and 16.64\% cumIoU. On DIOR-R-RSVG, it further achieves 67.23\% Pr@0.5, 56.73\% meanIoU, and 67.33\% cumIoU. These results demonstrate the effectiveness of O$^2$-VG-Trans for oriented visual grounding in remote sensing images.

\OtwoVGVLMResult
\OtwoVGUniRetrieval

\subsection{O$^2$-VG-Uni Results for Universal Oriented Proposals}

Since there are no existing universal proposal methods specifically designed for oriented objects, we adapt the widely used horizontal object proposal method RPN~\cite{faster_rcnn} to generate oriented proposals following the design of Oriented R-CNN~\cite{orientedrcnn}.

Table~\ref{Tab:OtwoVGUniRecall} presents the universal oriented proposal results with 100 and 300 proposals. Overall, O$^2$-VG-Uni achieves superior performance on VRSBench, AVVG, and DIOR-R-RSVG. In particular, with only 100 proposals, O$^2$-VG-Uni achieves AR$_{50:95}$ scores of 55.48, 73.61, and 59.76 on VRSBench, AVVG, and DIOR-R-RSVG, respectively.
When increasing the number of proposals to 300, O$^2$-VG-Uni further improves its recall across all datasets, achieving AR$_{50:95}$ scores of 56.41, 75.36, and 60.78, respectively.

In the first training stage, O$^2$-VG-Uni contains 44M learnable parameters. In the second stage, only the 512-dimensional objectness prompt is learnable, significantly reducing the number of trainable parameters.

Notably, O$^2$-VG-Uni adopts a real-time end-to-end Transformer architecture and runs at 83 FPS, faster than the Rotated RPN variants.

\subsection{O$^2$-VG-VLM Results for Visual Grounding}

We compare O$^2$-VG-VLM with GeoChat and GeoGround, two representative LLaVA-based methods~\cite{llava} that support oriented object visual grounding in remote sensing images. Both methods are fine-tuned following their official guidelines. As shown in Table~\ref{Tab:OtwoVGVLMResult}, O$^2$-VG-VLM consistently outperforms GeoChat and GeoGround across all three datasets and evaluation metrics, despite having fewer parameters. On VRSBench, it achieves 69.67\% Pr@0.5, 59.84\% meanIoU, and 59.36\% cumIoU. On AVVG, O$^2$-VG-VLM obtains 31.29\% Pr@0.5, 26.44\% meanIoU, and 27.52\% cumIoU. On DIOR-R-RSVG, it further achieves 78.35\% Pr@0.5, 67.85\% meanIoU, and 74.66\% cumIoU. These results demonstrate the effectiveness of O$^2$-VG-VLM for oriented visual grounding in remote sensing images.

\subsection{O$^2$-VG-Uni Results for Object Retrieval}

We further evaluate the object retrieval capability of O$^2$-VG-Uni, as shown in Table~\ref{Tab:OtwoVGUniRetrieval}. The precision, recall, and F1 scores are first computed within each class and then averaged across different classes. The threshold for CLIP and RemoteCLIP is based on cosine similarity, whereas that for O$^2$-VG-Uni is based on the prediction score.

Overall, O$^2$-VG-Uni demonstrates strong object retrieval capability across the three datasets. Specifically, it achieves F1 scores of 56.86\%, 3.65\%, and 49.25\% on VRSBench, AVVG, and DIOR-R-RSVG, respectively. Its recall reaches 53.11\%, 2.45\%, and 55.26\% on the three datasets, respectively.

\subsection{Ablation Study}

\OtwoVGTransEachModule
\OtwoVGUniEachModule
\OtwoVGVLMEachModule
\OtwoVLMMode

\textit{O$^2$-VG-Trans Ablation Study.}
Table~\ref{Tab:OtwoVGTransEachModule} presents an ablation study of O$^2$-VG-Trans on DIOR-R-RSVG. The full model achieves the best performance across all metrics, reaching 67.23\% Pr@0.5 and 56.73\% meanIoU. \ding{172} Removing encoder fusion causes the largest performance drop, reducing meanIoU by 4.89\%, which demonstrates the importance of early vision-language interaction. \ding{173} Replacing the proposed query selection with static query selection decreases meanIoU to 53.67\%, indicating the effectiveness of language-guided query selection. \ding{174} Removing text cross-attention also leads to a clear performance degradation, highlighting the importance of text-guided instance refinement. \ding{175} Replacing the proposed text prompt with a word-level prompt reduces the performance, validating the effectiveness of the proposed text representation. \ding{176} Using axis-aligned sampling points in deformable attention also degrades the performance, demonstrating the advantage of the proposed rotated deformable attention representation.

\OtwoVLMProposal
\otwovgvlmtrainloss

\textit{O$^2$-VG-Uni Ablation Study.}
Tab.~\ref{Tab:OtwoVGUniEachModule} conducts ablation studies on different components of O$^2$-VG-Uni on the DIOR-R-RSVG dataset.
\ding{172} Training all parameters in stage 2 without stage-1 training degrades the model into a binary oriented object detection framework, which achieves a high AR$_{50}$ of 89.84\% but completely fails in object retrieval. This indicates that although further fine-tuning improves the recall of universal oriented proposals, the alignment between visual and textual features is essential for object retrieval.
\ding{173} Training all parameters in stage 2 after stage-1 training further improves AR$_{50}$ to 94.62\%, but still fails to perform object retrieval. The reason is similar to that of \ding{172}.
\ding{174} Replacing the visual encoder from RemoteCLIP with CLIP decreases the performance. This demonstrates that a well-pretrained remote sensing visual encoder is better.
\ding{175} Using static query selection leads to decreases in result, verifying that adaptive query selection effectively removes redundant queries and improves the quality of reference oriented boxes.
\ding{176} Using axis-aligned sampling points in deformable attention results in performance degradation. This verifies that orientation-aware sampling is important.

\textit{O$^2$-VG-VLM Ablation Study.}
Table~\ref{Tab:OtwoVGVLMEachModule} presents the ablation study of different components in O$^2$-VG-VLM on the DIOR-R-RSVG dataset.
\ding{172} When the oriented box is directly tokenized as text, the performance drops significantly.
This is because continuous coordinate values may be split into multiple tokens by the tokenizer, which disrupts the spatial structure.
\ding{173} Replacing the $(cx,cy,w,h,\theta)$ representation with four vertex coordinates also causes a large performance degradation. This is mainly because the vertex-based representation requires predicting longer coordinate sequences (from 5 to 8 tokens) and is sensitive to vertex ordering, which may result in irregular quadrilateral outputs.
\ding{174} Removing universal oriented proposals reduces the meanIoU from 67.85\% to 61.90\%, showing that universal oriented proposals provide more accurate spatial priors.
Using only NTP (\ding{175}) or MTP (\ding{176}) limits the model's performance, indicating that the combination of NTP and MTP provides complementary supervision and enhances the oriented visual grounding capability.

\otwovgfamilyvisual

\otwovgtransresultvisual

\textit{O$^2$-VG-VLM Different Inference Modes.}
Table~\ref{Tab:OtwoVLMMode} compares the three inference modes of O$^2$-VG-VLM. The slow mode achieves the best performance on all three datasets due to reliable autoregressive decoding. The fast mode improves decoding efficiency but slightly reduces localization accuracy. The hybrid mode balances localization performance and inference efficiency. These results demonstrate the effectiveness of the proposed hybrid strategy.

\textit{Universal Oriented Proposals for O$^2$-VG-VLM.}
Table~\ref{Tab:OtwoVLMProposal} evaluates the effect of universal oriented proposals on O$^2$-VG-VLM. Introducing universal oriented proposals consistently improves localization performance across all three datasets. In particular, the improvements become more significant under higher IoU thresholds, indicating that the proposed proposals provide more accurate oriented object localization. For example, on DIOR-R-RSVG, the meanIoU improves from 61.90 to 67.85, while the cumIoU increases from 69.33 to 74.66. These results demonstrate that universal oriented proposals effectively enhance the grounding capability of O$^2$-VG-VLM.

Additionally, Fig.~\ref{fig:otwovgvlmtrainloss} presents the training loss curves of O$^2$-VG-VLM on the VRSBench, AVVG, and DIOR-R-RSVG datasets with and without universal oriented proposals. With universal oriented proposals, the training loss converges faster and achieves a more stable optimization process, demonstrating that high-quality oriented proposals facilitate more effective visual grounding learning.

\subsection{Qualitative Analysis}

\otwovguniresultvisual

\otwovguniretrievalvisual

\textit{Overall O$^2$-VG Family.}
Fig.~\ref{fig:otwovgfamilyvisual} provides an overview of the O$^2$-VG family on VRSBench. The figure presents the complete pipeline from referring expressions to oriented object localization. O$^2$-VG-Trans, as a discriminative Transformer-based model, directly predicts the oriented grounding box. O$^2$-VG-Uni extends O$^2$-VG-Trans to generate universal oriented proposals for potential foreground objects. These universal proposals are further incorporated into the input prompts of O$^2$-VG-VLM for vision-language visual grounding. The examples illustrate the different roles of the three models within the unified O$^2$-VG framework.

\textit{O$^2$-VG-Trans for Oriented Object Visual Grounding.}
Figs.~\ref{fig:otwovgtransresultvisual} shows qualitative results of O$^2$-VG-Trans on DIOR-R-RSVG and AVVG. The predicted oriented boxes closely follow the target objects under different object scales, orientations, and spatial distributions. The results also show that O$^2$-VG-Trans can capture detailed information in referring expressions, such as color attributes and relative positions.

\textit{O$^2$-VG-Uni for Universal Oriented Proposals.}
Fig.~\ref{fig:otwovguniresultvisual} visualizes the universal oriented proposals produced by O$^2$-VG-Uni on DIOR-R-RSVG, AVVG, and VRSBench. The proposals cover potential foreground objects using only an objectness prompt. These category-agnostic proposals provide object-level priors for subsequent O$^2$-VG-VLM.

\textit{O$^2$-VG-Uni for Object Retrieval.}
Fig.~\ref{fig:otwovguniretrievalvisual} presents the object retrieval results of O$^2$-VG-Uni on DIOR-R-RSVG. The retrieval heatmaps show that the proposal embeddings retain object-level semantic information. Proposals corresponding to the queried object receive higher similarity scores, enabling category-specific objects to be retrieved from the universal proposal set.

\textit{O$^2$-VG-VLM for Oriented Object Visual Grounding.}
Fig.~\ref{fig:otwovgvlmresultvisual} presents qualitative results of O$^2$-VG-VLM on DIOR-R-RSVG. The input prompt consists of the referring expression and universal oriented proposals. The predicted oriented boxes generally align well with the ground truth, demonstrating the ability of O$^2$-VG-VLM to perform oriented object visual grounding within an autoregressive vision-language framework.

\otwovgvlmresultvisual

\section{Conclusion}

In this paper, we propose the O$^2$-VG family for oriented object visual grounding in remote sensing images. O$^2$-VG-Trans introduces a cross-modality Transformer as a strong foundation. O$^2$-VG-Uni further provides universal oriented proposals using only an objectness prompt and supports efficient oriented object retrieval. Based on these proposals, O$^2$-VG-VLM adopts multi-token prediction in an autoregressive vision-language model for efficient oriented box generation. In addition, we construct DIOR-R-RSVG, a new dataset for oriented object visual grounding containing triplets of images, referring expressions, and oriented bounding boxes. Extensive experiments on multiple benchmarks demonstrate the effectiveness of the proposed unified framework.

\bibliographystyle{IEEEtran}
\bibliography{reference}

\end{document}